\documentclass[10pt,twocolumn,letterpaper]{article}

\usepackage{iccv}              

\usepackage{xcolor}         
\usepackage{makecell}

\usepackage{amsmath}
\usepackage{graphicx}
\usepackage{multirow}
\usepackage{colortbl}

\definecolor{iccvblue}{rgb}{0.21,0.49,0.74}
\usepackage[pagebackref,breaklinks,colorlinks,allcolors=iccvblue]{hyperref}

\def\paperID{7} 
\def\confName{ICCV}
\def\confYear{2025}

\title{DMS: Diffusion-Based Multi-Baseline Stereo Generation for Improving
Self-Supervised Depth Estimation}

\author{Zihua Liu$^{1}$\quad Yizhou Li$^{2}$\quad Songyan Zhang$^{3}$ \quad Masatoshi Okutomi$^{1}$ \\
$^{1}$Institute of Science Tokyo, Japan \\
$^{2}$Sony Semiconductor Solutions Group, Japan \\ 
$^{3}$Nanyang Technological University, Singapore \\ 
{\tt\small \{zliu$^{1}$,yli$^{2}$,mxo$^{1}$\}@ok.sc.e.titech.ac.jp, spyderzsy$^{3}$@gmail.com}
}

\begin{document}
\maketitle

\begin{abstract}
While supervised stereo matching and monocular depth estimation have advanced significantly with learning-based algorithms, self-supervised methods using stereo images as supervision signals have received relatively less focus and require further investigation. A primary challenge arises from ambiguity introduced during photometric reconstruction, particularly due to missing corresponding pixels in ill-posed regions of the target view, such as occlusions and out-of-frame areas. To address this and establish explicit photometric correspondences, we propose \textbf{DMS}, a model-agnostic approach that utilizes geometric priors from diffusion models to synthesize novel views along the epipolar direction, guided by directional prompts. Specifically, we finetune a Stable Diffusion model to simulate perspectives at key positions: left-left view shifted from the left camera, right-right view shifted from the right camera, along with an additional novel view between the left and right camera. These synthesized views supplement occluded pixels, enabling explicit photometric reconstruction. Our proposed DMS is a cost-free, ‘plug-and-play’ method that seamlessly enhances self-supervised stereo matching and monocular depth estimation, and relies solely on unlabeled 
stereo image pairs for both training and synthesizing. Extensive experiments demonstrate the effectiveness of our approach, with up to \textbf{35\%} outlier reduction and state-of-the-art performance across multiple benchmark datasets. The code is available at \url{https://github.com/Magicboomliu/DMS}.

\end{abstract}    
\section{Introduction} \label{introduction_section}

Depth estimation remains a critical task in computer vision with applications spanning autonomous driving\cite{stereodriving}, robotic navigation\cite{stereorobotic}, and augmented reality\cite{stereoar}. Although supervised stereo matching~\cite{ganet,IGEV-Stereo,gwcnet,GOAT,NiNet,psmnet,EDNet} and monocular depth estimation~\cite{DepthAnything,DepthAnythingV2,Metric3D,DepthPro,Zeo-Depth} have made substantial progress with deep learning, self-supervised methods using stereo images as supervision signals have become an appealing alternative, circumventing the costly need for labeled depth data. However, self-supervised methods face significant challenges, especially in occluded and out-of-frame regions where pixel correspondences are missed.

\begin{figure}[!t]
    \centering
\includegraphics[width=1.0\linewidth]{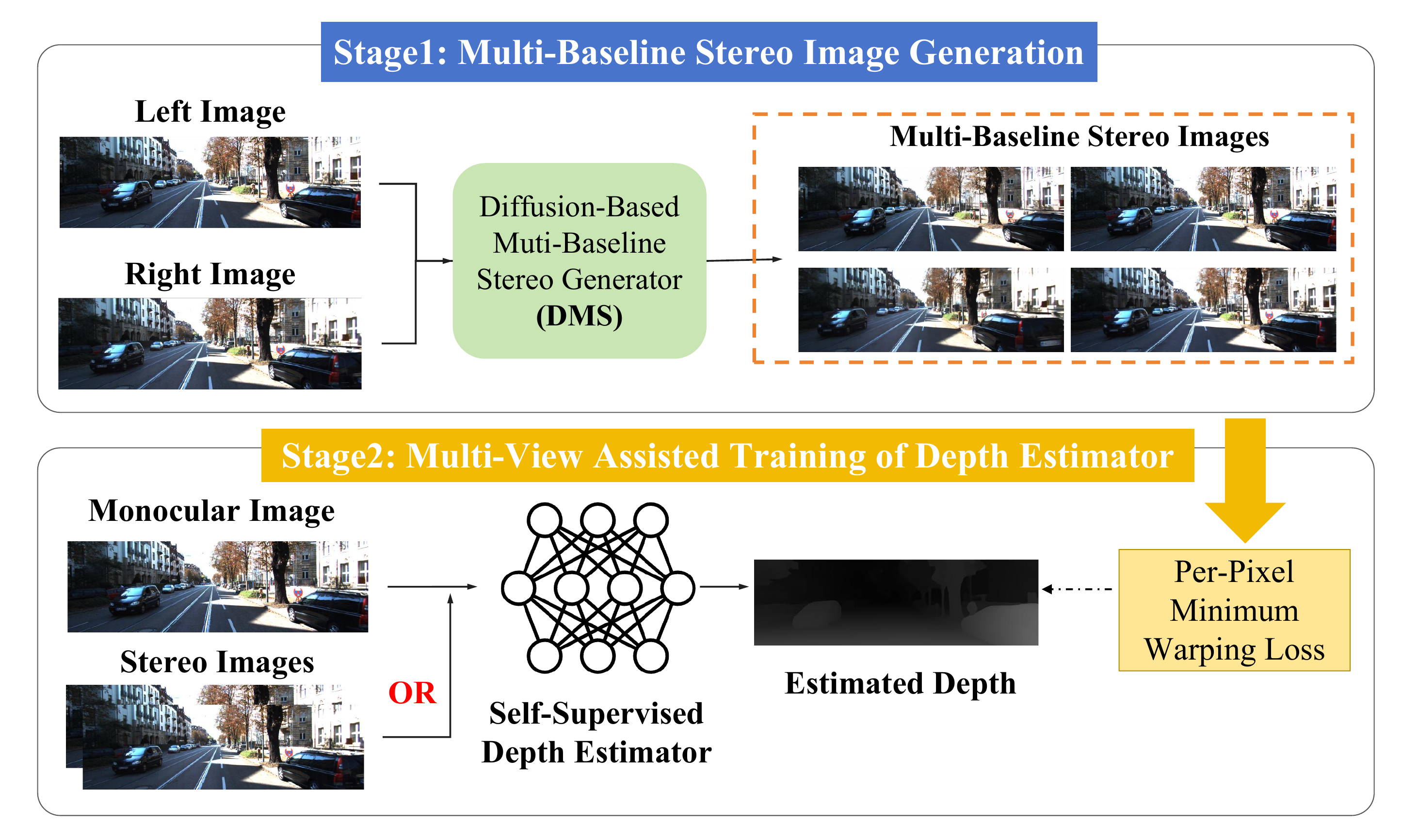}
    \caption{Overview of the proposed \textbf{D}iffusion-based \textbf{M}ulti-Baseline \textbf{S}tereo Generation (DMS) pipeline. Given the left and right input views, DMS generates additional multi-baseline views for the scene, which are then used to enhance self-supervised depth estimation through our Per-Pixel Minimum Warping Loss.}
\label{fig:overall_pipeline}
\end{figure}
Recent methods have attempted to mitigate the issues in these ill-posed regions by inferring or propagating context information~\cite{zhou2017unsupervised,OASM-Net,PASMNet,OASSNet,SegStereo,yu2016back}, or with the guidance of occlusion-aware modules~\cite{OASM-Net, PASMNet}, which have made notable improvements for self-supervised depth estimation using stereo images. However, heavy reliance on context propagation rather than direct matching leaves underlying reliability concerns unresolved. A promising alternative to improve the performance lies in extending the limited stereo viewpoints to a multi-baseline stereo setup. By introducing additional viewpoints along the epipolar line direction, occluded and ill-conditioned regions could potentially gain explicit matching pixels in the extended views. This prompts a critical question: \textit{can we generate reliable multi-baseline stereo images that enhance stereo pairs, enabling more accurate self-supervised depth estimation in these challenging regions?}

To address this, unlike other physical multi-camera systems \cite{EmbedMultiBaseline,Kanade1Okutomi,UMDEMB} with multiple baselines, in this paper, we aim to leverage the extensive priors from recent large-scale Latent Diffusion Models~(LDMs)~\cite{LatentDiffusion} and develop the \textbf{DMS~(Diffusion-based Multi-Baseline Stereo Generation)}, a model-agnostic method that uses geometric priors from LDMs to synthesize novel viewpoints and extend stereo baselines. During training, a Stable Diffusion model is conditioned on either the left or right stereo image and learns to generate the opposite view, guided by simple text prompts \textit{"to right"} or \textit{"to left"}. During inference, the learned DMS can be used to generate novel views (e.g., \textit{left-left} view by conditioning on the left image with the prompt \textit{"to left"} or \textit{right-right} view similarly). The proposed DMS effectively extends the baselines of the stereo images and enhances matching information for occluded and out-of-frame regions, thus significantly improving the disparity estimation in these regions. Our experiments indicate that the learned Stable Diffusion model inherently captures depth cues while learning transformations between left and right views, enabling it to generate reliable novel views. Additionally, we found that upscaling the input resolution yields the same pixel-level displacements in generated views, i.e., reduced displacements in original resolution. This allows precise control over shift distances with the scale factor and enables the generation of intermediate view, offering robust supervision signals for self-supervised depth estimation.
 
Our main contributions lie in three folds:

\begin{itemize} 
    \item We exploit geometric priors within the learned Stable Diffusion model for novel view synthesis along the epipolar line via direction prompts, relying solely on unlabeled stereo image pairs for both training and generation. 

    \item Our synthesized views fill in missing correspondences in ill-posed areas, including occluded regions and out-of-frame regions, providing explicit matching clues for stable photometric-based learning. 

    \item Our proposed \textbf{DMS} is a cost-effective, plug-and-play method that enhances self-supervised stereo matching and stereo-supervised monocular depth estimation, achieving state-of-the-art results with up to 35\% fewer outliers across multiple benchmarks. 

\end{itemize}
\section{Related Work}
\label{related_works}
\subsection{Self-Supervised Depth Estimation}
\label{self-supervised-depth-estimation}
\noindent\textbf{Self-Supervised Stereo Matching.} With limited access to densely labeled depth data, self-supervised stereo methods have been developed to learn disparity using photometric loss and left-right consistency checks~\cite{zhou2017unsupervised,OASM-Net,PASMNet,SegStereo,OASSNet,Flow2Stereo}. Early work showed that CNNs could learn stereo matching from photometric cues alone~\cite{zhou2017unsupervised}, while later methods added modules for handling occlusions~\cite{OASM-Net}, parallax attention to capture large disparities~\cite{PASMNet}, and semantic information to refine disparity~\cite{SegStereo}. Some methods also combine stereo and optical flow estimation to improve the disparity prediction~\cite{Flow2Stereo}.    

\noindent\textbf{Self-Supervised Monocular Depth Estimation with Stereo Supervision.} Self-supervised monocular depth estimation, which is normally trained with monocular video sequences aims to simultaneously estimate the camera poses and predict the scene depths. It has seen significant progress through the use of stereo image pairs, which provide a known relative pose, removing the need for a dedicated pose network like PoseNet. Monodepth~\cite{MonoDepth} introduced this approach by enforcing photometric consistency losses between stereo views, while Monodepth2~\cite{MonoDepth2} improved it with multi-scale depth predictions and auto-masking, achieving greater stability and precision. Building on these foundations, Watson et al.~\cite{watson2019self} integrated geometric hints to guide learning, and Guizilini et al.~\cite{guizilini20203d} employed 3D packing techniques to improve spatial detail and resolution. SDFANet~\cite{SDFANet} and DiffNet~\cite{DiffNet} each leverage feature-level information to refine depth predictions, achieving stronger internal consistency and handling of complex scene details.   

While current unsupervised depth estimation methods improve reliability in challenging areas such as occluded and out-of-frame regions, they still rely primarily on contextual similarity without explicit geometric cues. In contrast, our proposed DMS incorporates a plug-in tool that harnesses the powerful capabilities of a Stable Diffusion Model to enhance matching information in occluded regions through extended baseline images, significantly advancing disparity estimation where traditional methods encounter limitations.

\subsection{Multi-Baseline Stereo Image Synthesis}
\label{multi_view_synthesis}
\noindent\textbf{Geometry-Based Generation.} Previous deep learning-based approaches for novel view generation have primarily relied on geometry-based techniques such as adaptive convolution methods~\cite{xie2016deep3d,niklaus2017video,liu2018geometry,bello2020deep,bello2019deep}, which are often referred to as kernel estimation, where convolutional kernels dynamically adjust based on scene geometry.  Deep3D~\cite{xie2016deep3d} introduces a network that creates probabilistic disparity maps to blend shifted versions of the left-view image, synthesizing a right-view. Niklaus et al.~\cite{nichol2021glide} proposed adaptive separable convolutions (SepConv), approximating 2D convolutions with sequential vertical and horizontal 1D kernels for video frame interpolation. Deep3D pan~\cite{Deep3D_PAN} uses a novel T-shaped adaptive kernel with globally and locally adaptive dilation, integrating camera shift and local 3D geometry to synthesize natural 3D panned views from a 2D image. While these geometry-based methods can produce plausible multi-baseline stereo images, their quality remains low, particularly in challenging ill-posed areas.  

\noindent\textbf{Diffusion Models for Multi-View Synthesis.} Diffusion models \cite{ddpm, ddim, LatentDiffusion, controlnet} have excelled in 2D image generation. Works like DreamFusion \cite{poole2022dreamfusion} and SJC \cite{wang2023score} adapted 2D text-to-image models for 3D shape generation, spurring advancements in text-to-3D distillation \cite{chen2023fantasia3d,wang2023prolificdreamer,seo2023ditto,yu2023points,lin2023magic3d,seo2023let,tsalicoglou2023textmesh,zhu2023hifa,huang2023dreamtime,armandpour2023re,wu2023hd,chen2023it3d} based on score distillation sampling (SDS) loss. Additionally, several methods \cite{Zero123,watson2022novel,gu2023nerfdiff,deng2023nerdi,zhou2023sparsefusion,tseng2023consistent,chan2023generative,yu2023long,tewari2023diffusion,yoo2023dreamsparse,szymanowicz2023viewset,tang2023mvdiffusion,xiang20233d,liu2023deceptive,lei2022generative} use 2D diffusion models for multi-view image generation, reconstructing 3D scenes from single images with camera pose prompts. Other approaches \cite{tseng2023consistent,yu2023long, SyncDreamer, MVDiffusion, ERA3D} apply multi-view diffusion models, leveraging attention layers for image-conditioned novel view synthesis to ensure consistency across views. Most of these methods, however, require multi-view datasets (e.g., Objaverse~\cite{Objaverse}) for training. In contrast, our proposed DMS is trained solely on original stereo images from stereo datasets, eliminating the need for additional viewpoint information.

\section{Method}
\label{method}
In this section, we first provide an overview of latent diffusion models in Section~\ref{method:preliminaries}. We then present the details of our approach in subsequent sections. As illustrated in Figure~\ref{fig:overall_pipeline}, the DMS follows a two-stage training framework: multi-baseline stereoscopic view synthesis, followed by self-supervised depth estimation. In Stage 1, we train a Diffusion model to generate multi-baseline views from stereo images. In Stage 2, we use these generated views to train self-supervised depth networks, enhancing disparity estimation through multi-view consistency. Each stage is detailed in Sections~\ref{method:stage1_DMS} and Section~\ref{method:stage2_training}.

\subsection{Preliminaries}
\label{method:preliminaries}
\textbf{Latent Diffusion Model.}~The Latent Diffusion Model (LDM)~\cite{LatentDiffusion} substantially refines diffusion models by transitioning their operation into a latent space. This model utilizes an encoder to compress an image $x$ into a latent representation $z = E(x)$, effectively streamlining the learning of the distribution of latent codes, represented as $z_0 \sim p_{\text{data}}(z_0)$, consistent with the framework of Denoising Diffusion Probabilistic Models (DDPM) proposed by \cite{ddpm}. LDM operates through a bifurcated process: the forward process methodically adds Gaussian noise across time steps $t$ to form $z_t$, whereas the backward process focuses on noise reduction to more closely approximate the preceding, less noisy state $z_{t-1}$. The \textit{forward} process is described as:
\begin{equation}
    q(z_t|z_{t-1}) = \mathcal{N}(z_t; \sqrt{1 - \beta_t} z_{t-1}, \beta_t \mathbf{I}),
    \label{eq:forward_pass}
\end{equation}
where $\beta_t$ belonging to $\{\beta_{1},\beta_{2},...\beta_{T}\}$ is the variance scale of the forward process with $T$ steps. $\mathcal{N}$ represents the 
 gaussian distribution and $\mathbf{I}$ denotes the unit vector with the same size as the $z_t$. In the \textit{backward} process, the conditional denoising model strives to remove the noise and reconstruct the less noisy state $z_{t-1}$ as follows:
 \begin{equation}
p_\theta(z_{t-1}|z_t) = \mathcal{N}(z_{t-1}; \mu_\theta(z_t, t, \tau), \Sigma_\theta(z_t, t, \tau)),
\end{equation}
where $\tau$ denotes text embedding, $\mu_\theta$ and $\Sigma_\theta$ denote the mean and variance functions derived from the denoising model $\epsilon_\theta$, parameterized by $\theta$, which characterizes the current state's statistical properties. In the training phase, the model takes $z_t$ as the input, which is forwarded from $z_0$ by $t$ steps with sampled Gaussian noise $\epsilon$. The denoising U-Net estimates the noise $\hat{\epsilon}=\epsilon_\theta (z_{t},t)$, and the loss function $\mathcal{L}$ is expressed by:
\begin{equation}
\mathcal{L} = \mathbb{E}_{z_{0},\epsilon \sim \mathcal{N}(0,1), t \sim \mathcal{U}(T)} \left\|\epsilon-\hat{\epsilon}\right\|^{2}.
\end{equation}
Among the prominent LDMs, the Stable Diffusion \cite{LatentDiffusion} stands out as a prime example, trained extensively on large-scale image-text pair datasets, showcasing remarkable scalability across various tasks~\cite{Marigold,Zero123,SyncDreamer}.

\subsection{Diffusion-Based Multi-Baseline Stereo Generation (DMS).} 
\label{method:stage1_DMS} 

\begin{figure*}[!t]
    \centering
    \setlength{\abovecaptionskip}{0.1cm}
\includegraphics[width=1.0\linewidth]
{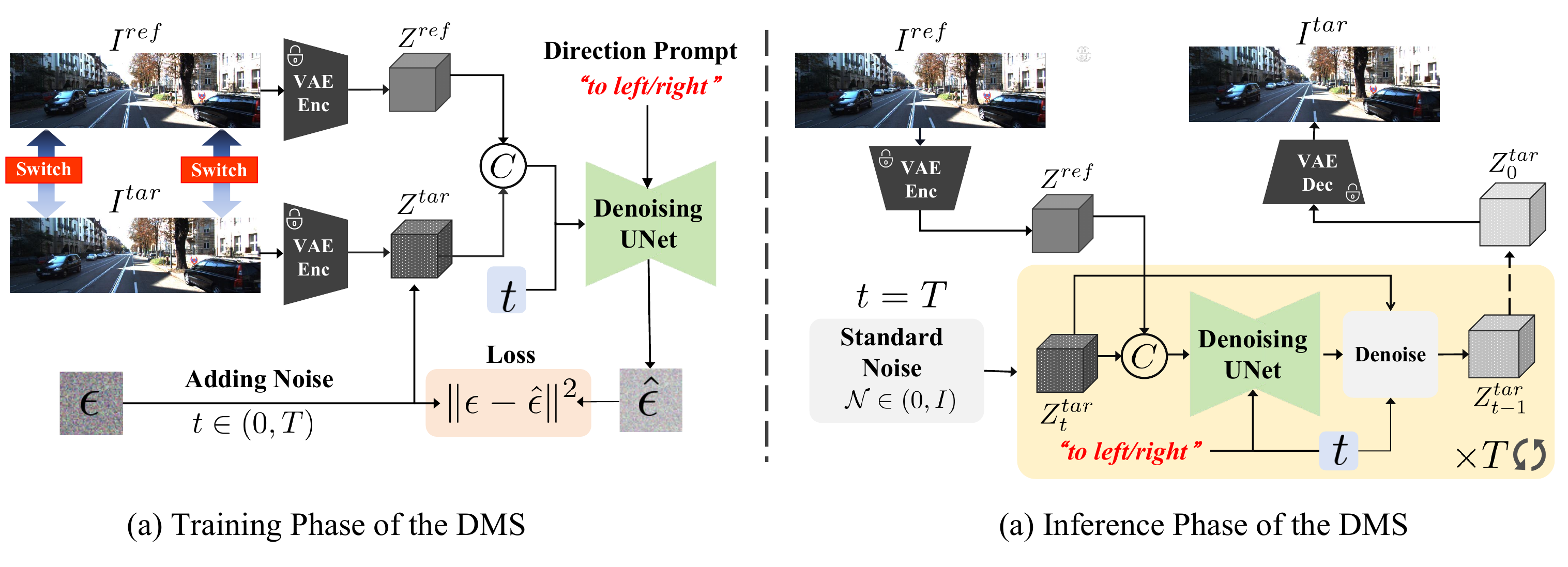}
    \caption{\textbf{Architecture of the proposed DMS.} (a) In the training phase, we fine-tune the Stable Diffusion model by conditioning it on a reference image (\textit{l} or \textit{r}) and a direction prompt specifying the transformation direction to reconstruct the opposite stereo view (\textit{r} or \textit{l}). (b) During inference, novel views are iteratively generated from standard Gaussian noise, guided by an input image and a direction prompt.}
\label{fig:DMS_Pipeline}
\end{figure*}

\subsubsection{Network Architecture}\label{network_architecture} 
The overall DMS pipeline is illustrated in Figure~\ref{fig:DMS_Pipeline}. Similar to \cite{Marigold}, we base our model on a pre-trained text-to-image LDM (Stable Diffusion V2), which has been learned from a large number of high-quality images from the  LAION-5B~\cite{LAION5B} dataset. We repurpose the Stable Diffusion model as an image-to-image generator by encoding the reference image as the image prompt, together with the direction text prompt to indicate the transformation direction.  
\subsubsection{Training Protocol: Direction Prompt Guided Fine-tuning with Stereo Images }\label{sec:training_protocal}
As shown in Figure~\ref{fig:DMS_Pipeline}.(a), we take the frozen VAE to encode both the reference view $I_{ref}$ and the target view $I_{tar}$ to form the latent features $Z_{ref}$ and $Z_{tar}$, respectively. During training, we set the $I_{ref}$ and $I_{tar}$ to be either the left or right images and switch step by step. 
To incorporate scene priors, we condition the denoising U-Net $\epsilon_{\theta}$ by concatenating the reference and target latents along the feature dimension, yielding $Z_{fus}^{t} = cat(Z_{ref}^{t}, Z_{tar}^{t})$. However, relying solely on a reference image is insufficient for determining the direction of transformation, leading to convergence difficulties of the diffusion model. To address this issue, we introduce text-based direction prompts to explicitly define the transformation direction. As shown in Table~\ref{tab:infernece_directions}, we utilize the pre-trained text encoder from the Stable Diffusion model to process simple prompts such as \textit{"to right"} and \textit{"to left"}. The resulting text embeddings serve as additional conditioning inputs for the denoising U-Net, enhancing the model’s representation capability. The repurposed denoising process can be described as follows:
\begin{equation}
    \epsilon = \epsilon_{\theta}(Z_{ref}^{t},Z_{tar}^{t}, t, \tau_{D}),
\end{equation}

\noindent here $\tau_{D}$ is the direction prompt indicts direction.

\begin{table}[!t]
\centering
\caption{Inference views with different direction text prompts}
\scalebox{0.8}{
\begin{tabular}{cl|cl|cl}
\hline
\multicolumn{2}{c|}{\textbf{Input View}} & \multicolumn{2}{c|}{\textbf{Direction Prompt}} & \textbf{Target View} \\ \hline
\multicolumn{2}{c|}{Left Image}           & \multicolumn{2}{c|}{\textit{to left}}                 & Left-Left Image (\textbf{Novel View})\\
\multicolumn{2}{c|}{Left Image}           & \multicolumn{2}{c|}{\textit{to right}}                & Right Image (Existing View)\\
\multicolumn{2}{c|}{Right Image}          & \multicolumn{2}{c|}{\textit{to left}}                 & Left Image (Existing View) \\
\multicolumn{2}{c|}{Right Image}          & \multicolumn{2}{c|}{\textit{to right}}                & Right-Right Image (\textbf{Novel View})   \\ \hline
\end{tabular}}
\end{table}
\label{tab:infernece_directions} 
\begin{figure}[!t]
    \centering
\includegraphics[width=1.0\linewidth]{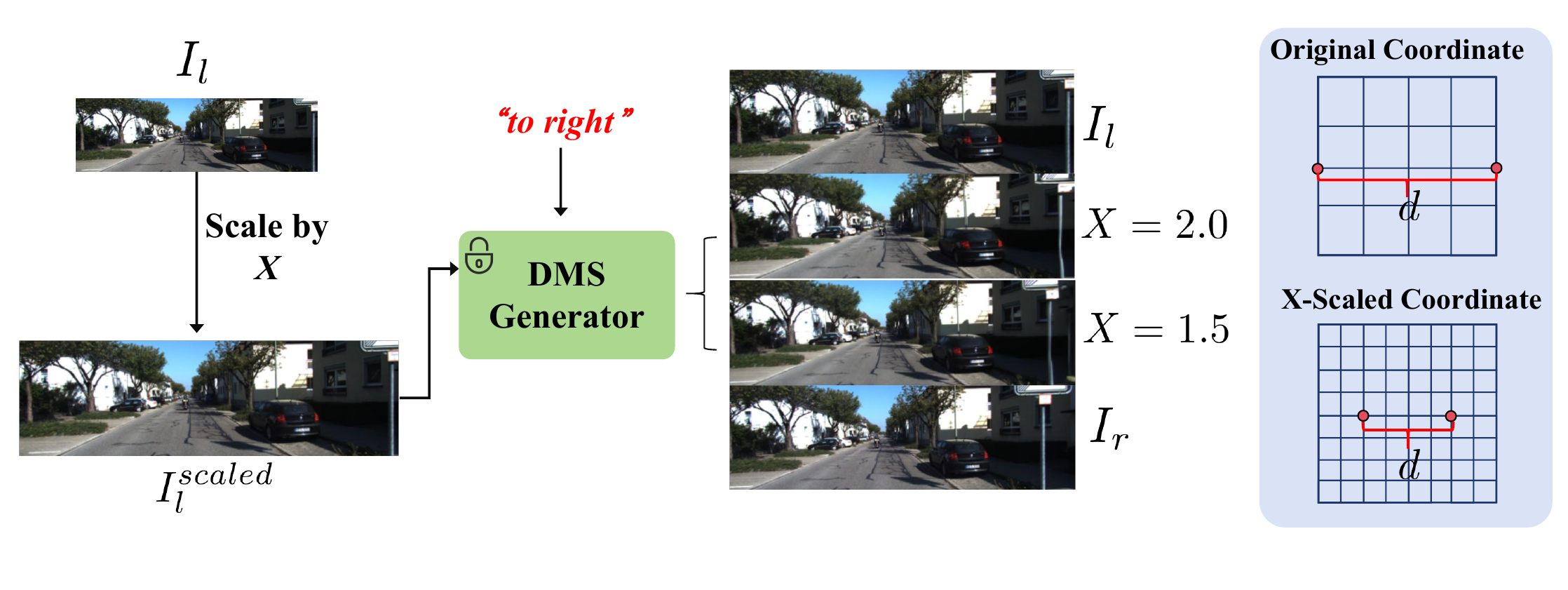}
    \vspace{-3mm}
    \caption{Intermediate view approximation with rescaling operation using pre-trained DMS generator.}
\label{fig:inference_med_view}
\end{figure}

\subsubsection{Inference: Multi-Baseline Stereo Generation}
\noindent \textbf{Extending Stereo Baseline with Direction Prompts.} After training the diffusion model, we utilize it to generate new views with input direction prompts. As shown in Table~\ref{tab:infernece_directions}, beyond the left and right views used for supervision, the fine-tuned diffusion model can generate additional \textit{left-left} and \textit{right-right} views by applying the appropriate direction prompt relative to the reference view. DMS achieves this by inputting the "to left" prompt alongside the left image, using the learned right-to-left transformation to generate a left-left view. Similarly, applying the reverse prompt enables right-right view synthesis. This technique effectively expands the stereo baseline without additional view supervision, providing a seamless approach to baseline extension.

\begin{figure}[!t]
    \centering
    \includegraphics[width=1.0\linewidth]{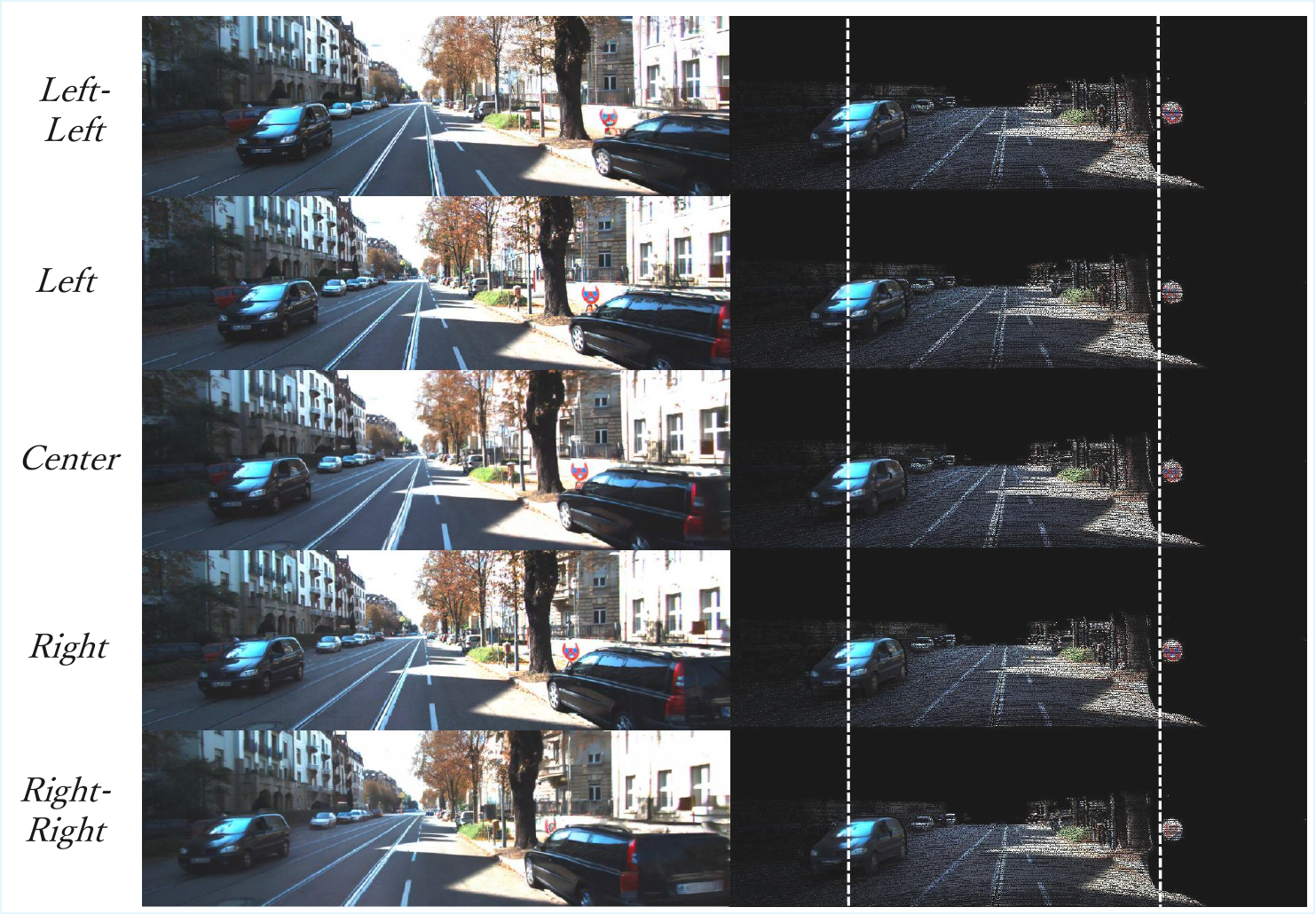}
    \caption{(a) Visualization of generated views on the KITTI dataset. From the top to the bottom: the \textit{left-left}, \textit{left}, \textit{center}, \textit{right}, and \textit{right-right} views. (b) Warped results from each view to the left image view using ground-truth sparse depth are depicted, with a white line marking the same horizontal location for reference.}
\label{fig:view_gen}
\end{figure}

\noindent \textbf{Intermediate View Approximation.} While the DMS enables the extension of the stereo baselines, they are limited to fixed baseline units. To enable finer disparity estimation, especially in occluded regions, we propose a simple yet effective method for generating intermediate perspectives with our fine-tuned diffusion model. As illustrated in Figure~\ref{fig:inference_med_view}, during the left-to-right generation with the prompt \textit{"to right"}, increasing the input image resolution by a factor $X>1$ creates an intermediate view between the left and right perspectives. The horizontal shift ratio follows an inverse relationship with the rescale factor $X$, indicating that our fine-tuned diffusion model inherently learns disparity cues through simple left-to-right and right-to-left transformations. Rescaling modifies the image coordinate system, resulting in intermediate views with proportionally smaller horizontal displacements, as shown in Figure~\ref{fig:inference_med_view}. Therefore, we leverage this property to generate intermediate views. In our experiments, we set 
$X=2.0$ to approximate a center view and rescale the output back to its original resolution. As shown in Figure~\ref{fig:view_gen}, we warp the generated views, including intermediate views, onto the left view using ground truth disparity. In Figure~\ref{fig:view_gen}. (b), the warped images align at the same horizontal positions, demonstrating the geometric consistency of the generated views. We further provide evaluation experiments on the synthesis dataset created by the CARLA~\cite{CARLA_Simulator} simulator to quantitatively assess the quality of the synthesized views. Please refer to the \textit{Supplementary Materials} for more details.

\subsection{Training of Self-Supervised Depth Estimators}
\label{method:stage2_training}
Upon acquiring multi-baseline images, we exploit these novel views to incorporate additional matching cues, thereby improving self-supervised depth estimators. Motivated by \cite{MonoDepth2}, we avoid averaging reprojection errors across source images when computing the loss, as it can mislead gradients in occluded regions. Instead, we adopt the \textit{Per-Pixel Minimum Warping Loss} \cite{MonoDepth2} for supervision. This approach is grounded in the straightforward hypothesis that warped source images using accurately predicted disparity or depth may not align with the target image in the occluded or out-of-frame regions, leading to significant photometric errors. In the stereo-matching setups, regions which are visible in the reference view and invisible (occluded) in its right view should be mostly visible in its opposite-side (left) view. By adopting the minimum of losses with multi-baseline warping from both sides, we effectively utilize matching clues from multi-baseline images to optimize disparity in ill-conditioned regions adaptively. The \textit{Per-Pixel Minimum Warping Loss} can be described as follows: 
\begin{gather}
\mathcal{L}_{\text{warp}} = \min_{i \in \mathcal{M}} \mathcal{L}_{pe}(I_l, \mathcal{W}(I_i, D \cdot s_i)), \\
\mathcal{L}_{pe}= \frac{\alpha}{2} (1 - \text{SSIM}(I_a, I_b)) + (1 - \alpha) \|I_a - I_b\|,
\end{gather} \label{ep:multi_view_eq}

\noindent where $\mathcal{M}$ is the scope of source images within the range of $\{r, ll,rr,c\}$, representing the \textit{right}, \textit{left-left}, \textit{right-right}, and \textit{center} view, respectively. $\mathcal{W}$ denotes the warping operation with disparity. $s_i$ are constant scale factors to scale the disparity for warping the source image to the target left image $\mathcal{I}_l$. Specifically, $s_r=1, s_{ll}=-1, s_{rr}=2, s_{c}=0.5$. The error metric $pe$ employs a hybrid loss combining L1 and SSIM \cite{SSIM_Loss}, following methodologies from \cite{zhao2016loss, MonoDepth} for calculating warping errors.

\section{Experiments}
\label{experiments}

\subsection{Implementation Details}\label{sec:imple}
\textbf{Datasets.} To demonstrate the versatility of our proposed DMS model, we trained the diffusion model on various stereo datasets: SceneFlow~\cite{Scenflow}, KITTI 2015~\cite{KITTI2015}, KITTI 2012~\cite{KITTI2012}, and MPI-Sintel~\cite{MPI-Sintel}. The SceneFlow dataset consists of 39,823 synthetic stereo pairs featuring flying objects with ground truth disparities. MPI-Sintel provides 1,064 synthesized stereo images with ground truth disparities. KITTI 2012 and 2015 provide real-world autonomous driving stereo images with sparse LiDAR-based disparities.

\noindent \textbf{Implementation Details of the Diffusion-Based Multi-Baseline Stereo Generator.} We implement the DMS using Pytorch with 4 NVIDIA A6000 GPUs and utilize the Stable Diffusion V2~\cite{LatentDiffusion} as the foundation model. During training, as outlined in Section~\ref{sec:training_protocal}, we froze the VAE and fine-tuned the Denosing UNet using Adam optimizer with a constant learning rate of $2e-5$ during all training phases with batch size 16.  We utilized the DDPM~\cite{ddpm} noise scheduler with a total of 1000 steps.  During inference, the DDPM sampling was configured to 32 steps for the KITTI dataset and 50 steps for the SceneFlow and MPI-Sintel datasets to ensure the generation quality.

\begin{table*}[!t]
\centering
\normalsize
 \setlength{\tabcolsep}{1.5mm}
 \renewcommand\arraystretch{1.2}

 \caption{Evaluation of generation quality on existing stereo perspectives across different datasets. We report the PSNR and SSIM of our generated left and right images via DMS. The "*" means using the CARLA simulator to test the intermediate views performance.}
 \scalebox{0.92}{
\begin{tabular}{c|cc|cc|ll|cc|ccccc}
\hline
\multirow{2}{*}{\textbf{Metric}} & \multicolumn{2}{c|}{SceneFlow~\cite{Scenflow}} & \multicolumn{2}{c|}{KITTI 2015~\cite{KITTI2015}} & \multicolumn{2}{l|}{KITTI 2012~\cite{KITTI2012}} & \multicolumn{2}{c|}{MPI-Sintel~\cite{MPI-Sintel}} & \multicolumn{5}{c}{CARLA*~\cite{CARLA_Simulator}} \\ \cline{2-14} 
 & Left & Right & Left & Right & Left & Right & Left & Right & Left-Left & Left & Center & Right & Right-Right \\ \hline
\textbf{PSNR}$\uparrow$ & 22.62 & 22.58 & 23.60 & 24.70 & 23.75 & 23.74 & 29.02 & 28.89 & 23.63 & 23.52 & 21.44 & 23.06 & 22.71 \\ \hline
\textbf{SSIM}$\uparrow$  & 0.73 & 0.73 & 0.80 & 0.81 & 0.74 & 0.75 & 0.86 & 0.86 & 0.76 & 0.76 & 0.72 & 0.75 & 0.72 \\ \hline
\end{tabular}
}
\label{tab:DMS_performance}
\end{table*}

\begin{table*}[!ht]
\centering
\normalsize
\caption{Quantitative evaluation of stereo image quality on the KITTI 2015 training set. Stereo Perceptual Consistency is measured by Stereo LPIPS (between left/right images) and Fusion SSIM (between grayscale-averaged cyclopean images). Geometric Consistency is assessed by EPE and D1 error from pretrained stereo networks, and photometric Warping Error under ground-truth disparities. \dag~ represents that using the official KITTI pre-trained weights.}
\scalebox{0.92}{
\begin{tabular}{c|cc|ccc}
\hline
\multirow{3}{*}{\textbf{Input Data}} & \multicolumn{2}{c|}{\textbf{Stereo Perceptual Consistency}} & \multicolumn{3}{c}{\textbf{Gemetric Consistency}} \\ \cline{2-6} 
 & \multicolumn{1}{c|}{\multirow{2}{*}{Stereo LPIPS $\downarrow$}} & \multirow{2}{*}{Fusion SSIM $\uparrow$} & \multicolumn{1}{c|}{\multirow{2}{*}{Warp Error$\downarrow$}} & \multicolumn{1}{c|}{IGEV-Stereo \dag\cite{IGEV-Stereo}} & NMRF-Stereo \dag \cite{NMRFStereo} \\ \cline{5-6} 
 & \multicolumn{1}{c|}{} &  & \multicolumn{1}{c|}{} & \multicolumn{1}{c|}{EPE$\downarrow$ /D1$\downarrow$} & EPE $\downarrow$/ D1$\downarrow$ \\ \hline
GT Left + GT Right & \multicolumn{1}{c|}{0.411} & 1.00 & \multicolumn{1}{c|}{0.123} & \multicolumn{1}{c|}{0.28 / 0.4} & 0.36 / 0.5 \\ \hline
\textbf{GT Left + DMS-Gen Right} & \multicolumn{1}{c|}{0.424} & 0.929 & \multicolumn{1}{c|}{0.138} & \multicolumn{1}{c|}{0.54 / 1.4} & 0.57 / 1.5 \\ \hline
\end{tabular}}
\label{exp:geometry_related_perception}
\end{table*}

\noindent \textbf{Implementation Details of the Self-Supervised Depth Networks.} The self-supervised stereo-matching networks are trained on the SceneFlow, KITTI, and MPI-Sintel datasets. For SceneFlow, we utilize the \textit{FlyingThings3D} subset with 19,984 images, excluding highly occluded scenes to ensure viable correspondences. Training is conducted with a batch size of 8 and a cosine decay learning rate starting from $1e^{-4}$. For MPI-Sintel, we follow a similar protocol, splitting the dataset for a $9:1$ evaluation ratio. On KITTI, we fine-tune the SceneFlow pre-trained model for 600 epochs at a resolution of 320$\times$960, with color augmentation as in \cite{AANet}. Please check \textit{Supplementary Material} for more details. For self-supervised monocular depth estimation with stereo-based supervision, the model is trained on KITTI-Raw and evaluated on the KITTI-eigen~\cite{eigen2014depth} split following the pipline of a previous study~\cite{MonoDepth2}.


\subsection{Evaluation of Generated Stereo Images}

\begin{figure}[!t]
    \centering
\includegraphics[width=1.0\linewidth]{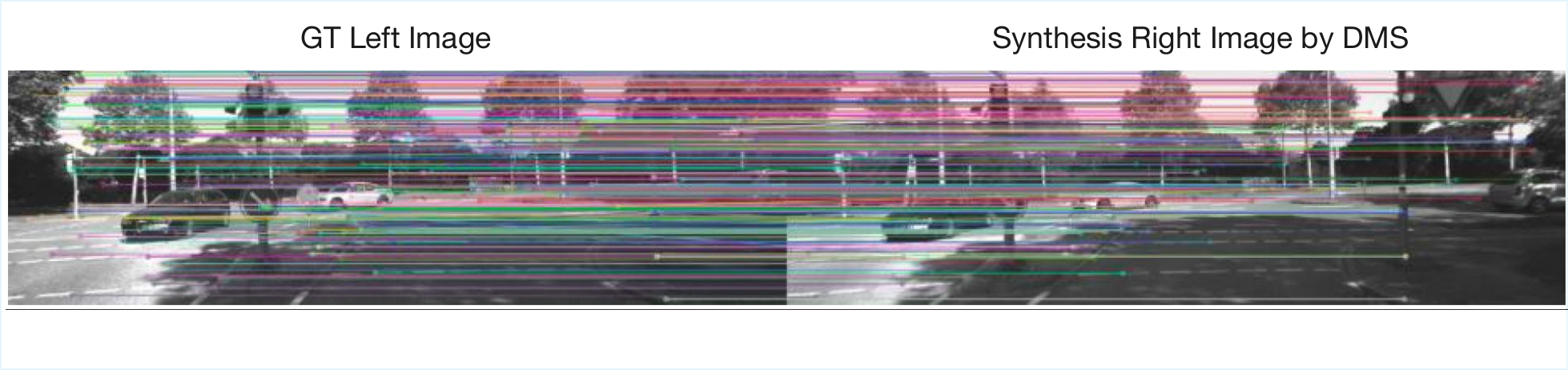}
    \caption{SIFT matches between GT left and DMS-synthesized right images predominantly display horizontal alignments.}
\label{fig:Epipolar_Check}
\end{figure}

As reported in Table~\ref{tab:DMS_performance}, we evaluate the PSNR and SSIM metrics to assess the quality of DMS-generated stereo views across different datasets. DMS achieves a PSNR of 22–23db on the synthetic SceneFlow dataset and a higher PSNR of 28–29db on the Sintel dataset. To further evaluate our synthesized stereo pairs, we adopt a dual-perspective evaluation inspired by prior work~\cite{TIP2015,TIP2017}, which highlights the importance of both binocular perceptual quality and geometric consistency. As the original metrics in~\cite{TIP2015,TIP2017} require non-public implementations and DMOS labels, we instead report simplified proxies.
 \textbf{(1) Stereo Perceptual Consistency:} We compute Stereo LPIPS (between left/right views) and Fusion SSIM (between cyclopean images) as surrogates for binocular realism.
\textbf{(2) Geometric Consistency}: We use pretrained stereo models~\cite{IGEV-Stereo,NMRFStereo} to estimate disparity from our synthesized stereo pairs, and report EPE, D1 Error, and Warp Error (under GT disparity).
As shown in Table~\ref{exp:geometry_related_perception}, our synthesized right images show strong perceptual alignment with the left view, achieving scores close to GT. The resulting disparity estimates exhibit only a minor drop in accuracy (NMRFStereo: EPE 0.36→0.57, IGEVStereo: 0.28→0.54), remaining within the 1-pixel tolerance commonly acceptable in practice. Furthermore, SIFT feature matching between ground truth and synthesized images, illustrated in Figure~\ref{fig:Epipolar_Check} , predominantly shows horizontal and parallel matches, validating the epipolar accuracy of our synthesized images.  

\begin{table}[!t]
\centering
\normalsize
\setlength{\tabcolsep}{1.0mm}
\renewcommand\arraystretch{1.2}
\caption{Comparisons of novel view synthesis quality on KITTI dataset with geometry-based multi-baseline synthesis methods. The 'K' denotes the KITTI dataset, and 'C' denotes the Cityscape dataset. \textbf{Bold} means the best performance.}
\scalebox{0.9}{
\begin{tabular}{c|c|c|c}
\hline
\textbf{Modeling} & \textbf{Training Dataset} & \textbf{PSNR} & \textbf{SSIM} \\ \hline
Deep3D~\cite{deep3d} & K & 20.07 & 0.637 \\ \hline
Deep3D-B~\cite{deep3d}  & K & 20.10 & 0.633 \\ \hline
SepConv~\cite{SepConv} & K & 19.73 & 0.633 \\ \hline
SepConv-D~\cite{SepConv} & K & 20.02 & 0.626 \\ \hline
monstet-net(Deep3D pan)~\cite{Deep3D_PAN} & K & 20.24 & 0.641 \\ \hline
monstet-net(Deep3D pan)~\cite{Deep3D_PAN} & K + CS & 20.76 & 0.677 \\ \hline
\rowcolor[rgb]{.949,.949,.949}
\textbf{DMS(Ours)} & K & \textbf{24.70} & \textbf{0.810} \\ \hline
\end{tabular}
}
\label{tab:dms_with_stage1_kitti}
\end{table}
\begin{table*}[!t]
\centering
\normalsize
\setlength{\tabcolsep}{1.0mm}
\renewcommand\arraystretch{1.0}
\caption{Ablation studies results on ScenceFlow, KITTI 2015, and MPI-Sintel datasets, employing PASMnet~\cite{PASMNet} as the baseline model. The terms \textit{ll}, \textit{rr}, and \textit{c} refer to the \textit{left-left}, \textit{right-right}, and \textit{center} views, respectively. Results include End-Point Error (EPE) and outlier ratios (errors $>3$px) across overall, occluded, and out-of-frame regions. \textbf{Bold} means the best performance.}
\scalebox{0.92}{
\begin{tabular}{c|cccccc|cccccc|cccccc}
\hline
\multirow{3}{*}{\textbf{Method}} & \multicolumn{6}{c|}{\textbf{SceneFlow}\cite{Scenflow}}                                                                           & \multicolumn{6}{c|}{\textbf{KITTI 2015~\cite{KITTI2015}}}                                 & \multicolumn{6}{c}{\textbf{MPI-Sintel~\cite{MPI-Sintel}}}                                                                          \\ \cline{2-19} 
& \multicolumn{3}{c|}{\textbf{EPE}$\downarrow$}                               & \multicolumn{3}{c|}{\textbf{\textgreater{}3px\%}$\downarrow$} & \multicolumn{3}{c|}{\textbf{EPE}$\downarrow$}                               & \multicolumn{3}{c|}{\textbf{\textgreater{}3px\%}$\downarrow$} & \multicolumn{3}{c|}{\textbf{EPE}$\downarrow$}                               & \multicolumn{3}{c}{\textbf{\textgreater{}3px\%}$\downarrow$} \\ \cline{2-19} 
& \textbf{All} & \textbf{Occ} & \multicolumn{1}{c|}{\textbf{Oof}} & \textbf{All}   & \textbf{Occ}   & \textbf{Oof}  & \textbf{All} & \textbf{Occ} & \multicolumn{1}{c|}{\textbf{Oof}} & \textbf{All}   & \textbf{Occ}   & \textbf{Oof}  & \textbf{All} & \textbf{Occ} & \multicolumn{1}{c|}{\textbf{Oof}} & \textbf{All}   & \textbf{Occ}  & \textbf{Oof}  \\ \hline
Baseline                         & 4.09         & 22.66        & \multicolumn{1}{c|}{10.83}        & 13.5        & 83.6         & 42.8        & 1.48         & 4.38         & \multicolumn{1}{c|}{9.26}         & 7.7          & 39.6         & 64.2        & 6.34         & 15.70        & \multicolumn{1}{c|}{14.27}        & 20.1         & 55.9        & 66.5        \\
+ \textit{ll} + \textit{rr}                        & 2.45         & 11.45        & \multicolumn{1}{c|}{6.16}         & 9.0          & 51.0         & 25.0        & 1.34         & 3.83         & \multicolumn{1}{c|}{7.82}         & 6.5          & 34.4         & 42.8        & 5.89         & 13.56        & \multicolumn{1}{c|}{11.92}        & 19.6         & 51.6       & 57.1        \\
+ \textit{c}     & 3.75         & 21.70        & \multicolumn{1}{c|}{7.73}         & 12.5         & 81.3         & 38.4        & 1.36         & 4.14         & \multicolumn{1}{c|}{7.64}         & 6.7          & 37.0         & 49.6        & 6.30         & 14.61        & \multicolumn{1}{c|}{\textbf{11.06}}        & 19.1         & \textbf{51.4}        & \textbf{52.9}        \\
+ \textit{ll} + \textit{rr} +\textit{c}                     & \textbf{2.32}         &\textbf{11.15}        & \multicolumn{1}{c|}{\textbf{5.57}}         & \textbf{8.4}         & \textbf{49.2}         & \textbf{23.4}        & \textbf{1.24}         & \textbf{3.56}         & \multicolumn{1}{c|}{\textbf{7.31}}         & \textbf{5.8}          & \textbf{32.3}         & \textbf{39.7}        & \textbf{5.53}         & \textbf{13.07}        & \multicolumn{1}{c|}{11.40}        & \textbf{18.9}         & \textbf{51.4}        & 53.0        \\ \hline
\end{tabular}
}
\label{tab:ablation_studies}
\end{table*}
\begin{table}[!t]
\centering
\normalsize
\setlength{\tabcolsep}{1.0mm}
\caption{ Benchmark results on KITTI 2015 test set. The rows highlighted in gray utilize our proposed DMS. The terms \textbf{"All"} represent overall regions, while \textbf{"fg"} and \textbf{"all"} denote foreground and overall regions, respectively.* indicates our re-implemented results after unsupervised training with vanilla warping loss and $\dag$ represents the result using the non-learning-based method where the employment of DMS utilizes multiple image inputs.}
\renewcommand\arraystretch{1.2}
\scalebox{0.82}{
\begin{tabular}{c|ccc}
\hline
\multicolumn{1}{c|}{} & \multicolumn{3}{c}{\textbf{All(\%) $\downarrow$}} \\ \cline{2-4} 
\multirow{-2}{*}{\textbf{Method}} & \textbf{D1-bg} & \textbf{D1-fg} & \textbf{D1-all} \\ \hline
\multicolumn{4}{c}{\textbf{Non-Learning Based Method}} \\ \hline
SGM$^\dag$~\cite{SGM} & 8.95 & 20.55 & 10.88 \\ 
\rowcolor[rgb]{.949,.949,.949} SGM$^\dag$ + DMS & 7.96(11.1\% $\downarrow$) & 16.68(18.8\%$\downarrow$) & 9.41(13.5\%$\downarrow$) \\ \hline
\multicolumn{4}{c}{\textbf{Learning-Based Methods}} \\ \hline
Zhou \textit{et.al}~\cite{zhou2017unsupervised} & - & - & 9.91 \\
SegStereo~\cite{SegStereo} & - & - & 8.79 \\
OASM\cite{OASM-Net} & 6.89 & 19.42 & 8.98 \\
Flow2Stereo\cite{Flow2Stereo} & 5.01 & 14.62 & 6.61 \\ 
\hline
PASMnet\cite{PASMNet} & 5.41 & 16.36 & 7.23 \\
\rowcolor[rgb]{.949,.949,.949} PASMnet + DMS & 5.24(3.1\%$\downarrow$) & 13.96(14.7\%$\downarrow$) & 6.69(7.5\%$\downarrow$) \\ 
StereoNet*\cite{khamis2018stereonet} & 7.31 & 17.77 & 9.05 \\
\rowcolor[rgb]{.949,.949,.949} StereoNet* + DMS & 4.68(36.0\%$\downarrow$) & 12.06(32.1\%$\downarrow$) & 5.91(34.7\% $\downarrow$)\\ 
CFNet*~\cite{shen2021cfnet} & 7.22 & 18.54 & 9.11 \\
\rowcolor[rgb]{.949,.949,.949} CFNet* + DMS & 4.64(35.7\%$\downarrow$) & 12.33(33.5\%$\downarrow$) & 5.92(35.0\%$\downarrow$) \\ 
RaftStereo*~\cite{lipson2021raft} & 3.38 & 13.62 & 5.08 \\
\rowcolor[rgb]{.949,.949,.949} RaftStereo* + DMS & 2.95(12.7\%$\downarrow$) & 6.88(50.9\%$\downarrow$) & 3.60(29.1\%$\downarrow$) \\ 
IGEVStereo*~\cite{IGEV-Stereo} & 3.76 & 11.14 & 4.98 \\
\rowcolor[rgb]{.949,.949,.949} IGEVStereo* + DMS & 2.80(25.5\%$\downarrow$) & 6.37(45.27\%$\downarrow$) & 3.40(31.72\%$\downarrow$) \\ 

{MCStereo*~\cite{feng2024mc}}& 3.01 & 13.38 & 4.73 \\

 \rowcolor[rgb]{.949,.949,.949} {MCStereo* + DMS} & 2.83(6.0\%$\downarrow$) & 7.03(47.46\%$\downarrow$) & 3.67(22.41\%$\downarrow$) \\
 \hline
\end{tabular}
}
\vspace{-4mm}
\label{tab:improving_stereo}
\end{table}

Besides, we compare DMS's performance on the KITTI dataset against state-of-the-art geometry-based multi-baseline synthesis methods, including  Deep3D~\cite{deep3d}, Deep3D Pan~\cite{Deep3D_PAN}, and SepConv~\cite{SepConv}. Table~\ref{tab:dms_with_stage1_kitti} shows that DMS, trained solely on KITTI Raw, surpasses Deep3D Pan (trained on KITTI and Cityscapes) by approximately 4db in PSNR and 0.133 in SSIM, underscoring DMS’s superior quality in stereoscopic image synthesis. Additionally, we evaluate DMS performance under ill-posed conditions, with detailed results provided in the \textit{Supplementary Materials}.

\begin{table*}[!t]
\centering
\normalsize
\setlength{\tabcolsep}{2.0mm}
\caption{Evaluation results on KITTI Eigen test set. The rows highlighted in gray represent the methods that utilize the proposed DMS. "*" represents a slight modification to model implementation. Refer to \textit{Supplementary Material} for more details. }

\scalebox{0.95}{
\begin{tabular}{c|cccc|ccc}
\hline
\textbf{Method} & \textbf{Abs Rel$\downarrow$} & \textbf{Sq Rel$\downarrow$} & \textbf{RMSE$\downarrow$} & \textbf{RMSE\_log$\downarrow$} & \textbf{A1$\uparrow$} & \textbf{A2$\uparrow$} & \textbf{A3$\uparrow$} \\ \hline
MonoDepth~\cite{MonoDepth} & 0.120 & 1.041 & 5.272 & 0.217 & 0.849 & 0.944 & 0.974 \\
\rowcolor[rgb]{.949,.949,.949} MonoDepth + DMS & \textbf{0.109} & \textbf{0.860} & \textbf{5.004} & \textbf{0.202} & \textbf{0.865} & \textbf{0.952} & \textbf{0.978} \\ \hline
MonoDepth2~\cite{MonoDepth2} & 0.109 & 0.873 & 4.960 & 0.209 & 0.864 & 0.948 & 0.975 \\
\rowcolor[rgb]{.949,.949,.949} MonoDepth2 + DMS & \textbf{0.105} & \textbf{0.811} & \textbf{4.850} & \textbf{0.200} & \textbf{0.873} & \textbf{0.963} & \textbf{0.988} \\ \hline
SDFANet*~\cite{SDFANet} & 0.104 & 0.997 & 4.583 & 0.186 & 0.888 & 0.962 & 0.981 \\
 \rowcolor[rgb]{.949,.949,.949} SDFANet* + DMS & \textbf{0.097} & \textbf{0.643} & \textbf{4.218} & \textbf{0.181} & \textbf{0.891} & \textbf{0.963} & \textbf{0.983} \\ \hline
DiffNet~\cite{DiffNet} & 0.104 & 0.809 & 4.766 & 0.201 & 0.879 & 0.953 & 0.976 \\ \hline
 \rowcolor[rgb]{.949,.949,.949}  DiffNet + DMS~\cite{DiffNet} & \textbf{0.098} & \textbf{0.726} & \textbf{4.606} & \textbf{0.191} & \textbf{0.886} & \textbf{0.958} & \textbf{0.980} \\ \hline
 
\end{tabular}}
\label{tab:monocular_improvement_tab}
\end{table*}


\subsection{Self-Supervised Stereo Matching}
\subsubsection{Ablation Studies}
To validate the effectiveness of the newly generated views by the proposed DMS on self-supervised stereo matching, we carried out ablation studies across various datasets, with results presented in Table~\ref{tab:ablation_studies}. We report the End-Point Error (EPE) and outlier ratios (errors$>$3px) for overall, occluded, and out-of-frame regions to illustrate the efficacy of our method. The occlusion and the out-of-frame masks, which are essential for evaluations, were generated using a left-right consistency check, detailed further in our \textit{Supplementary Material}. For the evaluation of the KITTI 2015 \& 2012 dataset, we follow the same strategy used in \cite{zhou2017unsupervised}, where we randomly pick 160 image pairs from training set for training, and the other 40 image pairs for validation.

\begin{figure}[!h]
    \centering
\includegraphics[width=1.0\linewidth]{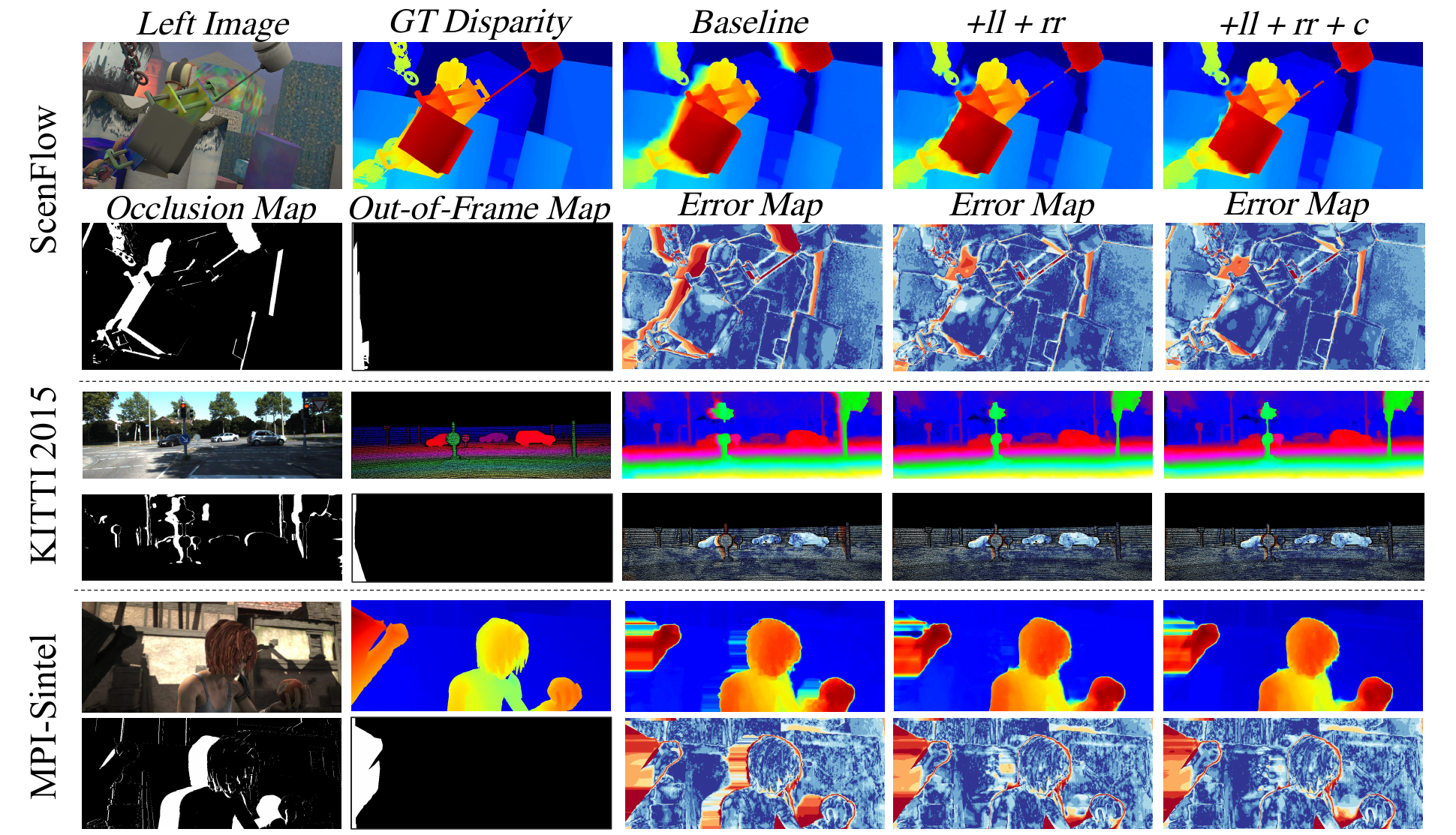}
    \caption{Ablation results on SceneFlow, KITTI 2015, and MPI-Sintel. For each dataset, the first row shows estimated disparities; the second row shows error maps and ill-conditioned regions (e.g., occlusions, out-of-frame areas).}
\label{fig:ablations_comparsion}
\vspace{-4mm}

\end{figure}
\begin{figure}[!t]
    \centering
\includegraphics[width=1.0\linewidth]{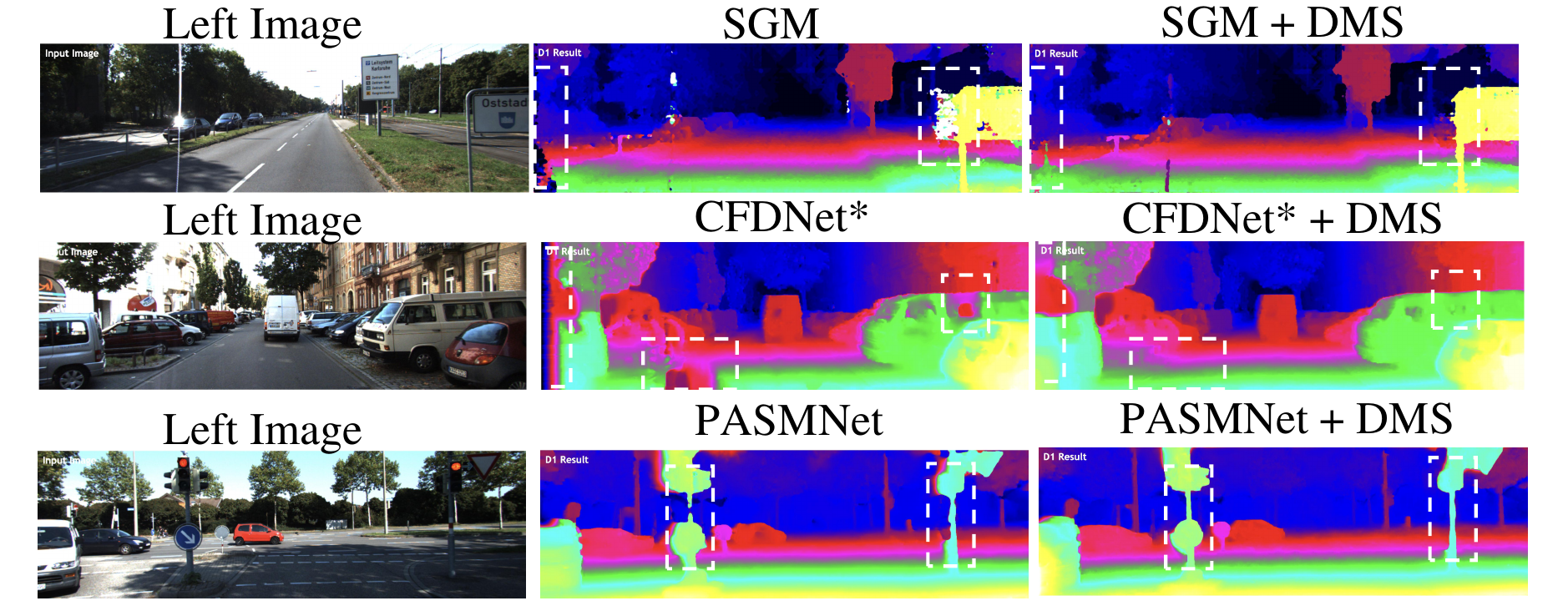}
    \caption{Visualization comparison on KITTI 2015 test set. The proposed DMS significantly improves the disparity estimation at the occluded regions and borders. Zoom in for a better view.}
    \vspace{-4mm}
\label{fig:improving_unsup_stereo}
\end{figure}

\noindent Table~\ref{tab:ablation_studies} illustrates that leveraging novel views generated by the proposed DMS for loss computation markedly improves disparity estimation without modifying the network's structure, evidenced by improved EPE and D1 values compared to the Baselines. With the proposed DMS, the model exploits geometric consistency across multi-baseline views and get optimized by the multi-view consistency warping loss. This notably improves disparities in occluded regions, with enhancements of 50.8\%, 18.7\%, and 17.2\% on SceneFlow, KITTI, and Sintel datasets, respectively. The extended views (\textit{left-left}\&\textit{right-right}) generated by the DMS also significantly boost out-of-frame disparity estimation. Figure~\ref{fig:ablations_comparsion} visually compares these enhancements across datasets, highlighting the Baseline model's limitations in reconstructing structured disparities, particularly in occluded areas. In contrast, the multi-baseline images from the DMS model enable the model to produce a more accurate reconstruction of structured disparities with reduced errors, underlining the efficacy of our approach in improving geometric precision for stereo-matching tasks.
\begin{figure}[!t]
    \centering
    \includegraphics[width=0.95\linewidth]{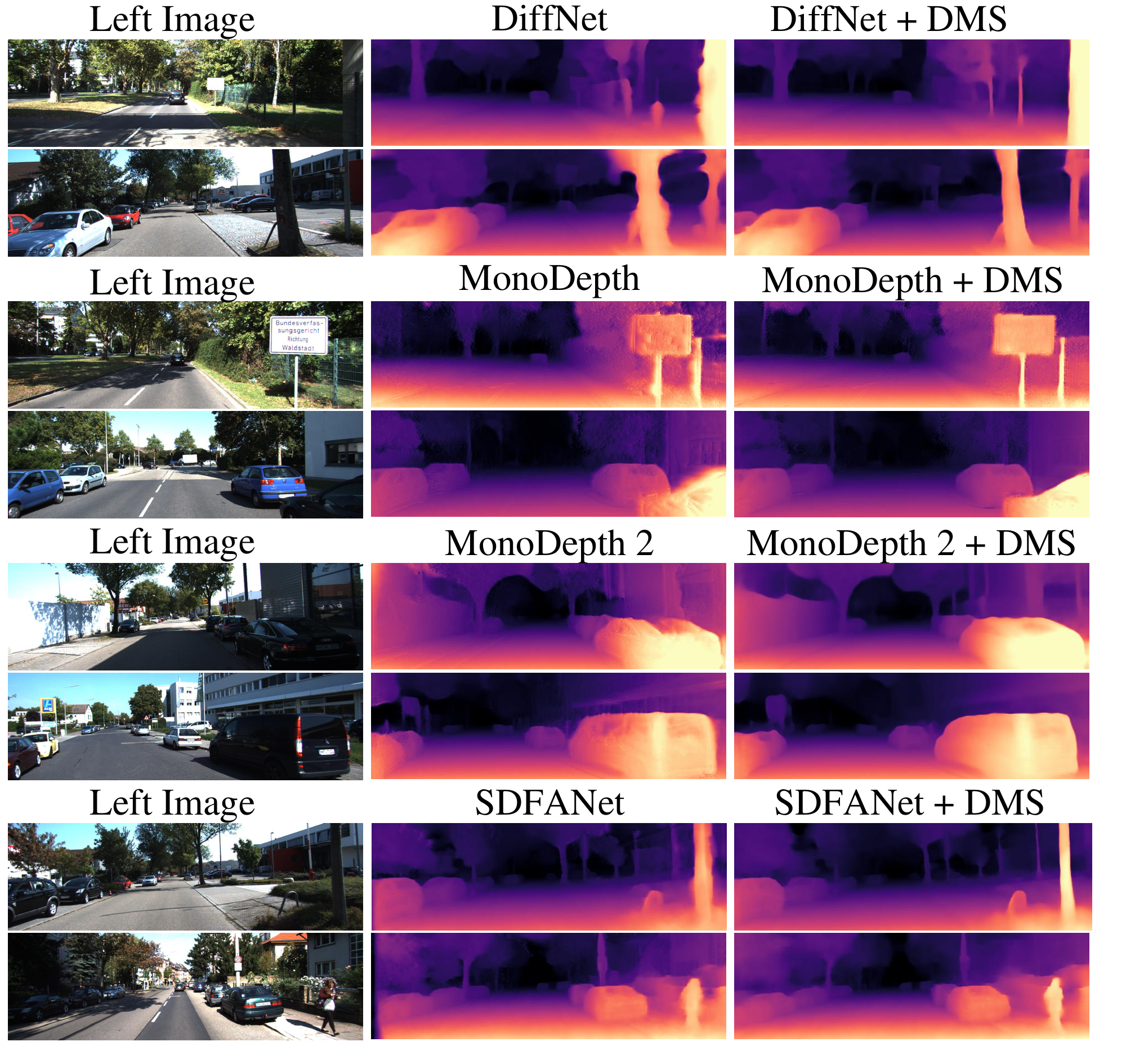}
    \caption{Visualization comparison of self-supervised monocular depth estimations across various existing methods.}
\label{fig:supp:monocular_evaluation}
\vspace{-4mm}
\end{figure}

\subsubsection{Evaluation on Self-Supervised Stereo Matching.}
Our Diffusion-based Multi-baseline Stereo Generator (DMS) functions as a versatile "plug-in" that leverages existing unlabeled stereo images to train and generate multi-baseline views, allowing for seamless integration into any stereo-matching framework. We further evaluated DMS's flexibility across various stereo matching networks on the KITTI testing set by fine-tuning a pre-trained SceneFlow model on a mixed KITTI 2012 and 2015 dataset over 600 epochs. For non-learning-based Semi-Global Matching (SGM)~\cite{SGM}, we integrate DMS by adapting the cost volume construction with multi-baseline information as detailed in Equation~\ref{ep:multi_view_eq}. As shown in Table~\ref{tab:improving_stereo}, the integration into SGM led to substantial improvements, particularly for foreground objects, with a 16.1\% improvement in non-occluded regions and an even greater 18.8\% improvement across all regions, suggesting high effectiveness in occluded areas though not explicitly evaluated. For learning-based methods, DMS integration significantly enhances performance, with most methods following SGM's trend and demonstrating its effectiveness in both non-occluded and occluded regions. Additionally, it can further boost the state-of-the-art supervised methods including RAFTStereo~\cite{lipson2021raft} and IGEVStereo~\cite{IGEV-Stereo} and MCStereo~\cite{feng2024mc} for self-supervised fine-tuning on the KITTI dataset. Figure~\ref{fig:improving_unsup_stereo} illustrates the improvements of our DMS pipeline for both traditional and learning-based stereo matching methods. Incorporating multi-baseline images leads to significant enhancements in challenging regions, such as occluded and out-of-frame areas, highlighted by white bounding boxes.


\subsection{Evaluation on Self-Supervised Monocular Depth Estimation}
\label{sec:exp:improve_self_supervised_monocular}
The primary advantage of our proposed method is its dual capability: the generated multi-baseline image views are useful not only for self-supervised stereo matching but also for improving self-supervised monocular depth estimation. To evaluate this, we extend our DMS framework to monocular depth estimation on the KITTI raw dataset. Following the standard protocol established in \cite{MonoDepth2}, we assess performance on the KITTI Eigen test set. Table~\ref{tab:monocular_improvement_tab} and Figure~\ref{fig:supp:monocular_evaluation} demonstrate the effectiveness of integrating the DMS pipeline into existing methods, showing consistent performance improvements across different network architectures. This underscores the broad applicability of DMS in enhancing self-supervised monocular depth estimation.

\section{Conclusions}
\label{conclusions}

In this paper, we have presented DMS, a cost-free model-agnostic approach that exploits geometric priors from large-scale Diffusion models to synthesize novel views across different baselines, thereby supplementing occluded pixels for explicit photometric matching. The proposed DMS synthesizes novel perspectives for the left-left and the right-right baseline camera, along with an intermediate novel view within the original field of vision. Extensive experiments demonstrate that DMS can significantly enhance self-supervised stereo matching and stereo-based monocular depth estimation in a 'plug-and-play' manner. The DMS allows leveraging unlabeled stereo images to enhance depth estimation without additional annotations, providing a promising avenue for further advancements in the field.

\clearpage
\setcounter{page}{1}
\maketitlesupplementary
\label{appendix}
In this supplementary material, we provide additional details and results to complement the main paper. Section~\ref{sec:supp:additional_details} outlines the experimental setup, including dataset configurations and hyperparameters. Section~\ref{sec:supp:additional_experimental_results} provides more experimental results both in the image quality of the DMS and subsequent training of the self-supervised stereo-matching networks. Finally, in Section~\ref{sec:supp:additional_vis_results} we illustrate more multi-baseline stereo image results across different datasets using the proposed DMS.
\subsection{Additional Implementation Details.}\label{sec:supp:additional_details}
\subsubsection{Additional Implementation Details of Diffusion-Based Multi-Baseline Stereo Generator (DMS).}
\label{sec:supp:DMS}
In this section, we detail the training process of the Diffusion-based Multi-baseline Stereo Generator (DMS) across various datasets to ensure reproducibility. We implement the DMS using Pytorch with the diffusers~\cite{diffusers} as the code base and utilizing the Stable DiffusionV2~\cite{LatentDiffusion} as the initial parameter weight.  We further report the computation resources that are needed in the inference stage in Table~\ref{tab:computer_resource}.

\noindent\textbf{SceneFlow.~}For training on the SceneFlow dataset, we employ the widely-used FlyingThings3D test set \cite{PASMNet, OASM-Net, AANet, ganet, GOAT, IGEV-Stereo, NiNet} and a training subset of 19,984 images from the FlyingThings3D set, filtering out scenes where occlusion exceeds 80\%. To conform to the input size requirements of the Stable Diffusion Model, which necessitates divisibility by 8, the original images are resized from $540 \times 960$ to $576 \times 960$ using top padding. We optimize memory usage by employing a batch size of 1 with a gradient accumulation equivalent to a batch size of 16. Optimization is performed using the Adam optimizer with a constant learning rate of $2e-5$ under half-precision (float16) settings. The training spans 20 epochs, with the resultant model used for both evaluation and generating new views. Inference utilizes a DDPM scheduler with a step of 50 denoising process for view synthesis. The efficacy of the generated views is quantitatively assessed in Table \ref{tab:DMS_performance}.  

\noindent\textbf{KITTI 2015~\&~2012.} For the limited view of the KITTI~2015 and 2012 datasets, we fine-tune the Diffusion model on the KITTI raw dataset~\cite{KITTIRaw} which compromises over 43,482 stereo images, where we split the 400 images and 394 images containing in KITTI 2015 and KITTI 2012 dataset and use the left views for training. We pad the original resolution of $375 \times 1242$ and $374 \times 1238$ into $284 \times 1248$ to meet the input size requirements of the Stable Diffusion Model. We use the same optimizer and learning rate that is adopted in training the ScenceFlow model and training for 10 epochs to get the final model. Inference utilizes a DDPM scheduler with a step of 32 denoising process for view synthesis. For further used in unsupervised stereo matching, we fine-tune the KITTI-raw pre-trained model on KITTI~2012 and KITTI~2015 datasets, respectively. Note that we both generate views on the KITTI raw dataset for improvement in the performance of monocular depth estimators as outlined in Section~\ref{sec:exp:improve_self_supervised_monocular}.  

\noindent\textbf{MPI-Sintel Dataset.} For the MPI-Sintel dataset, we partition the dataset into a training set and an evaluation set using a $9:1$ ratio. The training utilizes the "final pass" images, and we adjust the original resolution from $436 \times 1024$ to $440 \times 1024$ to accommodate the model's input requirements. The optimization parameters, including the optimizer and learning rate, are consistent with those used for the SceneFlow model. The training duration is set to 50 epochs to finalize the model. During inference, view synthesis is performed using a DDPM scheduler with a 50-step denoising process.

\noindent\textbf{CARLA Dataset.} Existing stereo datasets typically contain only \textit{left} and \textit{right} views, making it challenging to evaluate the quality of extended multi-baseline images. To address this, we utilize the CARLA simulator \cite{CARLA_Simulator} to generate a synthetic multi-baseline stereo dataset with 1000 image pairs (\textit{left}, \textit{center}, \textit{right}, \textit{left-left}, \textit{right-right}) under 15 diverse weather conditions (e.g., \textit{ClearNoon}, \textit{WetNight}). The dataset is split into training and testing sets with a 9:1 ratio. For fine-tuning DMS, we adopt the same training protocol as the KITTI dataset, training for 50 epochs with KITTI pre-trained weights as initialization. During inference, we also utilize a DDPM scheduler with a step of 32 denoising processes for view synthesis. The performance of the generated multi-baseline images can be outlined in Table~\ref{tab:novel_view_of_Carla}.

\subsubsection{Additional Implementation Details of Training the Self-Supervised Depth Estimators.} In this section, we provide a detailed description of the implementation details used for training self-supervised depth networks with multi-baseline images generated by DMS. This includes ablation studies settings and experiments across different datasets on both self-supervised stereo matching and monocular depth estimation.

\noindent\textbf{Implementation Details of the Ablation Studies.} We leveraged PASMNet~\cite{PASMNet} as our baseline to evaluate the effectiveness of our multi-baseline stereo images in improving disparity estimation in self-supervised stereo matching settings. We conduct ablation studies on two NVIDIA 3090 GPUs with PyTorch. For the SceneFlow dataset, training was performed on the FlyThings3D subset (as detailed in Section~\ref{sec:supp:DMS}) and evaluated on the SceneFlow official test set.  The model was trained with a batch size of 8, a disparity range of 192, and 100,000 steps. The initial learning rate was set to $1 \times 10^{-4}$ and reduced using cosine decay. The checkpoint with the lowest End-Point Error (EPE) on the validation set was selected for final evaluation, the results of which are presented in Table~\ref{tab:ablation_studies}. Fine-tuning on the KITTI dataset followed the protocol in \cite{zhou2017unsupervised}, using 160 image pairs for training and 40 for validation, with weights pre-trained on SceneFlow. The training procedure was consistent with SceneFlow, incorporating data augmentation techniques from \cite{AANet}, such as random cropping and adjustments to brightness, saturation, and contrast. For the Sintel-MPI dataset, the model was trained from scratch using the same parameters as SceneFlow to ensure methodological consistency. This approach validates the robustness of our method across diverse datasets.

\noindent\textbf{Implementation Details of Training the Self-Supervised Stereo Matching Networks.} Besides the SceneFlow dataset, we further test the performance of the DMS integrated into existing stereo-matching networks to validate the 'plug-in-and-play' ability on the KITTI dataset. For the KITTI 2015 benchmark, we deployed the model which was initially pre-trained on the SceneFlow dataset and subsequently fine-tuned on a combined dataset of KITTI 2012 and 2015, encompassing 394 images. We selected the model with the optimal D1 value for submission to the official KITTI 2015 benchmark to obtain our final results. Considering the limited availability of open-source self-supervised stereo-matching methods, we extended the applicability of our proposed DMS by adapting supervised networks like RaftStereo~\cite{lipson2021raft} and IGEVStereo~\cite{IGEV-Stereo} to self-supervised settings using photometric warping loss, demonstrating the method's versatility and broad potential for adaptation.  

\noindent\textbf{Implementation Details of Training the Self-Supervised Monocular Depth Estimators.}  All models were trained on the full KITTI Eigen training set (45,200 stereo image pairs) and validated on a small set (4,424 images). We selected models with the lowest validation error and tested them on the KITTI Eigen test set (697 images). For SDFANet \cite{SDFANet}, marked as * in the main paper, we made small changes to its original loss computation to incorporate our proposed DMS properly during training. SDFANet predicts a disparity cost volume with shape $B\times C\times D^*\times H\times W$, where $D^*$ is the number of disparity candidates. Disparity is estimated using Soft Argmin on the third dimension. The loss is computed by warping the left image with all disparity candidates, warping each sub-cost of one disparity candidate with shape $B\times C\times H\times W$ using the corresponding disparity, and calculating a weighted sum to synthesize the right image. The loss is then computed using the synthesized right image and the input right image. To match other methods and incorporate the proposed DMS, we used the same warping loss as the other three compared methods. We warped the right image to the left using the estimated disparity from SDFANet and computed the loss with the input left image. Similarly, by incorporating DMS, the disparity is used to warp the left-left image, right-right image, and center image to compute additional losses.

\subsubsection{Evaluation Details of the Self-Supervised Depth Estimations.}
\noindent\textbf{Occlusion and Out-of-Frame Mask Generation for Evaluation.} The MPI-Sintel dataset provides the ground truth occlusion and the out-of-frame mask for evaluation, but SceneFlow and KITTI did not provide such specific masks for evaluation. To address this issue, we use the same strategy used in ~\cite{GOAT} by using the left-right consistency to generate the occlusion mask and the ground disparity to calculate the out-of-frame mask. The process can be described as follows: 
\begin{equation}
M_{occ}=
\begin{cases}
1& \text{if $D_{\Delta}(x,y)\geq1,$}\\
0& \text{otherwise,}
\end{cases} 
\end{equation}

\begin{equation}
M_{oof}=
\begin{cases}
1& \text{if $D_{shift}(x,y)<0,$}\\
0& \text{otherwise,}
\end{cases} 
\end{equation}
\begin{align}
D_{\Delta}(x,y) = \lvert d_{l}(x,y) - d_{r}(x+d_{l}(x,y),y) \rvert , \\
D_{shift}(x,y) = x - d_{l}(x,y),
\end{align}
where $M_{occ}$ and $M_{off}$ represents the generated occlusion masks and the out-of-frame mask, respectively. And the $d_{l}$ and $d_{r}$ are the ground truth disparity map. For the reason that the KITTI dataset only provides the ground-truth sparse disparity maps for the left images, which makes it difficult to directly apply the left-right consistency check to generate the occlusion masks, following the strategy utilized in \cite{GOAT}, we use a pre-trained model~\cite{LeaStereo} to generate the pseudo-left and pseudo-left disparities, the left-right consistency check between the pseudo disparity maps for the left view and the right view is applied to generate a pseudo occlusion mask for performance evaluation.

\noindent\textbf{Evaluation Metrics for Self-Supervised Stereo Matching.} To showcase the effectiveness of our proposed DMS, especially on ill-conditioned regions, we report the End-Point-Error~(EPE) and the $>3$px outliers(percentage of the error bigger than 3 pixels) on overall regions, the occluded regions, and out-of-frame regions, respectively. The definition of the EPE is as follows: 
\begin{align}
EPE(d,\hat{d})= \lvert d-\hat{d} \rvert.
\end{align}
For the performance on the KITTI 2015 validation set, we report the $>3px$ which describes the outliner ratio of the predicted disparity. where can be described as follows:
\begin{align}
>3px = \frac{N_{\Delta{e}>3px}}{N_{total}}, \quad \Delta{e}= \lvert d-\hat{d} \rvert,
\end{align}
\noindent where $N$ means the number of pixels.

\noindent For the KITTI 2015 testing benchmark, we follow the official evaluation protocol to report the D1-value as shown in Table~\ref{tab:improving_stereo}.

\noindent\textbf{Evaluation Metrics for Self-Supervised Monocular Depth Estimation.} 
We evaluate each method using several metrics from prior work \cite{eigen2014depth}, which uses the predicted depth $d^*$ and GT depth $\hat{d}^*$ in meters to compute the errors:
\begin{gather}
    AbsRel = \frac{1}{|T|}\sum_{d^*\in T}\frac{|d^* - \hat{d}^*|}{\hat{d}^*}, \\
    SqRel = \frac{1}{|T|}\sum_{d^*\in T}\frac{||d^* - \hat{d}^*||^2}{\hat{d}^*}, \\
    RMSE = \sqrt{\frac{1}{|T|}\sum_{d^*\in T}||d^* - \hat{d}^*||^2}, \\
    RMSE(\log) = \sqrt{\frac{1}{|T|}\sum_{d^*\in T}||\log d^* - \log \hat{d}^*||^2}, \\
 A(thr) = \max(\frac{d_i^*}{\hat{d}_i^*},\frac{\hat{d}_i^*}{d_i^*}) = \delta < thr ,
\end{gather}

\noindent where $T$ denotes all the test pixels in all test image samples, and A1, A2, A3 denote the $thr$ be set as $1.25$, $1.25^2$, and $1.25^3$ respectively.

\begin{table}[!t]
\centering
\setlength{\tabcolsep}{1.0mm}
\renewcommand\arraystretch{1.0}
\caption{Computation resources for utilizing the DMS to generate multi-baseline stereo images across different datasets with different resolutions. Note the inference time and GPU Memory are tested on a single NVIDIA A6000 GPU.}
\scalebox{0.88}{
\begin{tabular}{c|c|c|c}
\hline
\textbf{Dataset} & \textbf{Denoising Steps} & \textbf{\begin{tabular}[c]{@{}c@{}}Inference Time\\ Per Image\end{tabular}} & \textbf{GPU Memory} \\ \hline
SceneFlow~\cite{Scenflow} & 50 & 5.34 s & 7.61 G \\ \hline
KITTI~\cite{KITTI2012} & 32 & 4.12 s & 6.82 G\\ \hline
MPI-Sintel~\cite{MPI-Sintel} & 50 & 6.04 s & 6.72 G \\ \hline
CARLA~\cite{CARLA_Simulator} & 32 & 4.30 s & 7.05G \\ \hline
\end{tabular}}

\label{tab:computer_resource}
\end{table}

\subsection{Addition Experimental Results.}
\label{sec:supp:additional_experimental_results}
In addition to the experimental results presented in the main paper, this section provides supplementary evaluations to thoroughly demonstrate the validity of our proposed Diffusion-Based Multi-Baseline Stereo Generation (DMS) and its impact on improving the performance of self-supervised depth estimation methods.
\subsubsection{Diffusion-Based Multi-Baseline Stereo Generation }

\noindent\textbf{More Ablations on Rescale-Factor \textit{X}}. We select a rescale-factor of 2.0 in the paper as the \textit{center} view provides the most effective representation of the intermediate view between left and right images. To further justify this choice, we extend the ablation studies in Table ~\ref{ab:more_ablations_X} with additional candidates (0.5, 1.5, 3.0), as shown in Table.\textcolor{red}{1} below. These factors produce denser intermediate views, such as \(\frac{1}{3}\) \textit{l} \(\rightarrow\) \textit{r} and \(\frac{2}{3}\) \textit{l} \(\rightarrow\) \textit{r}, improving performance over the baseline. However, the most effective configurations remain the default left-left (\textit{+ll}), right-right (\textit{+rr}), and center view (\textit{+c}), highlighted in \textcolor{gray}{gray}, which cover most out-of-view and occluded regions. Moreover, applying all new views together further improves overall EPE to 1.22, closely matching the 1.24 achieved with only \textit{+ll+rr+c}. Considering both performance and efficiency, we chose a rescale-factor of 2.0 for intermediate view generation in the paper.
\begin{table}[!t]
\centering
\normalsize
\setlength{\tabcolsep}{1.0mm}
\renewcommand\arraystretch{1.0}
\caption{Additional ablations on the KITTI2015 validation set with varying rescale factors. *Generate left-left and right-right images with a 0.5 scale. }

\begin{tabular}{c|c|ccc|lll}
\hline
\multirow{2}{*}{\textbf{\begin{tabular}[c]{@{}c@{}}Factor\\ X\end{tabular}}} & \multirow{2}{*}{\textbf{\begin{tabular}[c]{@{}c@{}}Views\\ Used\end{tabular}}} & \multicolumn{3}{c|}{\textbf{EPE\(\downarrow\)}} & \multicolumn{3}{c}{\textbf{D1\(\downarrow\)}} \\ \cline{3-8} 
 &  & \multicolumn{1}{l}{\textbf{All}} & \multicolumn{1}{l}{\textbf{Occ}} & \multicolumn{1}{l|}{\textbf{Oof}} & \textbf{All} & \textbf{Occ} & \textbf{Oof} \\ \hline
- & \textit{l,r} & 1.48 & 4.38 & 9.26 & 7.7 & 39.6 & 64.2 \\
0.5 & \textit{+ ll *+rr*} & 1.37 & 3.97 & 7.90 & 6.8 & 36.1 & 48.0 \\
\rowcolor[rgb]{.949,.949,.949}
1.0 & \textit{+ ll+rr} & 1.34 & 3.83 & 7.82 & 6.5 & 34.4 & 42.8 \\
1.5 & +\(\frac{2}{3}\) \textit{l} \(\rightarrow\) \textit{r}  & 1.40 & 4.22 & 8.17 & 7.1 & 38.1 & 51.7 \\
\rowcolor[rgb]{.949,.949,.949}
2.0 & +\textit{c} & 1.36 & 4.14 & 7.64 & 6.7 & 37.0 & 49.6 \\
3.0 & +\(\frac{1}{3}\) \textit{l} \(\rightarrow\) \textit{r} & 1.41 & 4.16 & 8.33 & 7.1 & 36.7 & 52.9 \\ \hline
All Above & + \textit{all} & 1.22 & 3.44 & 7.20 & 5.6 & 31.7 & 39.3 \\ \hline
\end{tabular}
\label{ab:more_ablations_X}
\end{table}

\noindent\textbf{Computation Resources Analysis}. We report the computation resources for utilizing the DMS to generate multi-baseline stereo images using an NVIDIA A6000 GPU and Intel i9-13900KF CPU. The image resolutions for inference are 540 $\times$ 960, 384 $\times$ 1280, 436 $\times$ 1024, and 540 $\times$ 960 for SceneFlow, KITTI, MPI-Sintel, and CARLA datasets, respectively. This demonstrates that our DMS can efficiently perform inference on a single GPU with less than 8GB of memory, highlighting its practical applicability.

\begin{table*}[!t]
\centering
\caption{Novel view quality evaluations on synthesis dataset created by CARLA~\cite{CARLA_Simulator} simulator. We report the PSNR and SSIM for both the left view, right view, left-left view, right-right view, and center view, respectively.}
\setlength{\tabcolsep}{1.2mm}
\renewcommand\arraystretch{1.2}
\begin{tabular}{c|c|c|c|c|c|c}
\hline
\textbf{Generated View} & \textbf{Input View} & \textbf{Direction Prompt} & \textbf{Upscaling} & \textbf{Novel View} & \textbf{PSNR} & \textbf{SSIM} \\ \hline
Left & Right & \textit{to left} & - &  & 23.52 & 0.76 \\
Right & Left & \textit{to right} & - &  & 23.06 & 0.75 \\
Left-Left & Left & \textit{to left} & - & \checkmark & 23.63 & 0.76 \\
Right-Right & Right & \textit{to right} & - & \checkmark & 22.71 & 0.72 \\
Center & Left & \textit{to right} & $\times$2 & \checkmark & 21.44 & 0.72 \\ \hline
\end{tabular}
\label{tab:novel_view_of_Carla}
\end{table*}
\begin{figure}[!t]
    \centering
    \includegraphics[width=1.0 \linewidth]{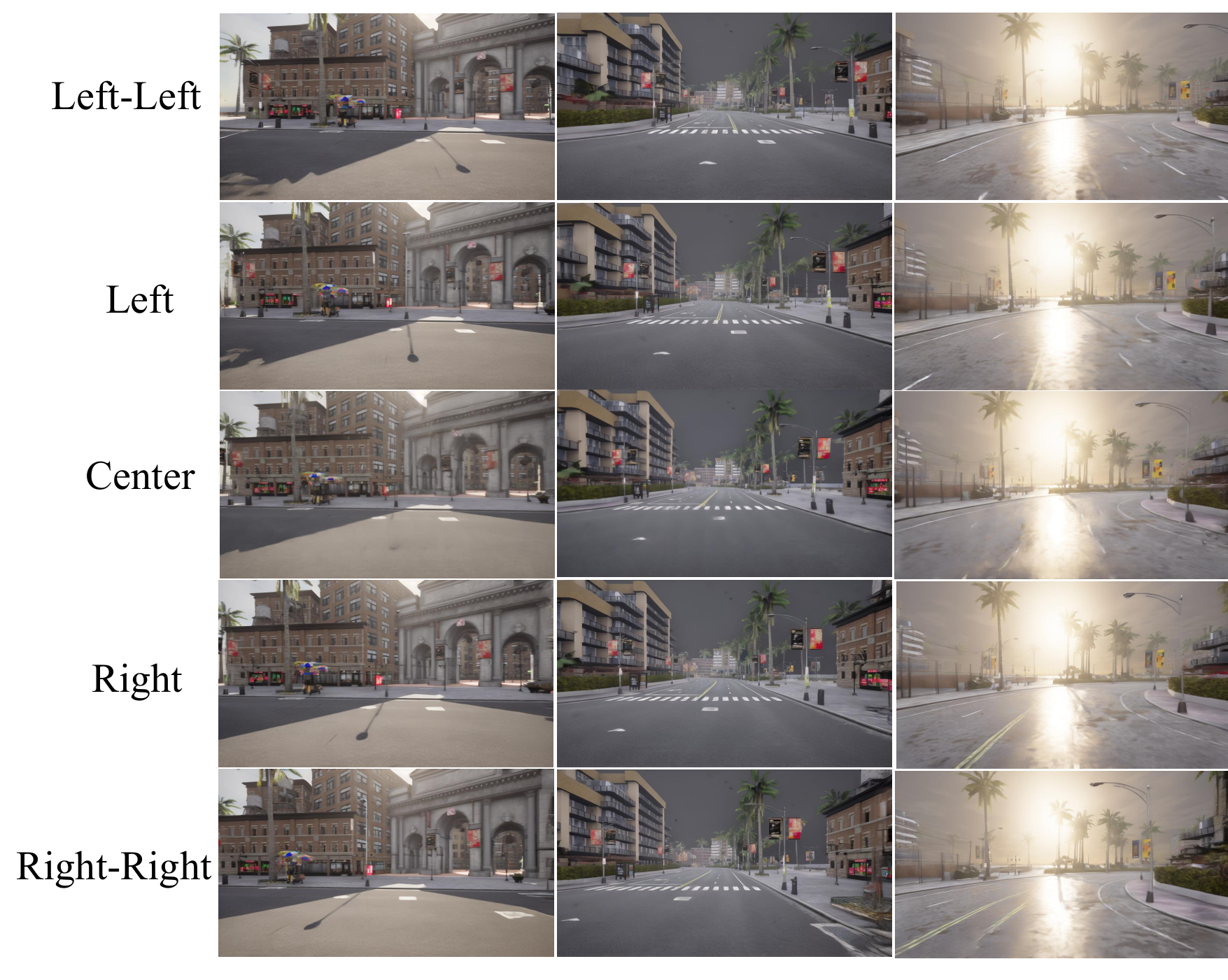}
    \caption{Multi-baseline stereo images generation using proposed DMS on the CARLA synthesis dataset.}
\label{fig:supp:mpi_CARLA}  
\end{figure}
\noindent\textbf{Multi-Baseline Stereo Image Evaluation on CARLA.} Stereo datasets typically provide only \textit{left} and \textit{right} views, limiting the evaluation of extended multi-baseline images. To address this, we generate a synthetic multi-baseline stereo dataset using the CARLA simulator \cite{CARLA_Simulator}, consisting of 1000 image pairs (\textit{left}, \textit{center}, \textit{right}, \textit{left-left}, \textit{right-right}) across 15 weather conditions (e.g., \textit{ClearNoon}, \textit{WetNight}). During training, we also only used the left and right view to train the DMS and used the pre-trained DMS model to generate multi-baseline stereo images. As shown in Table~\ref{tab:novel_view_of_Carla}, we report the PSNR and SSIM of the generated views including the novel view left-left, right-right, and center view, respectively. The check-marked annotations indicate the newly generated perspectives (left-left and right-right) obtained using our proposed inference method. These views exhibit the comparable performance of PSNR and SSIM to the rendered left and right views, despite the absence of ground truth left-left and right-right views during training. While the generated intermediate views show a slight decrease in PSNR, their SSIM remains consistent with other views. This demonstrates that the multi-baseline images produced by DMS maintain geometric consistency, making them valuable for improving self-supervised depth estimation. Further visualization results are illustrated in Section~\ref{sec:additional_visualziation_results} in this supplementary material.

\begin{table}[!t]
\centering
\setlength{\tabcolsep}{1.5mm}
\caption{Ablation Studies On KITTI 2012 dataset for self-supervised streo matching.The terms \textit{ll}, \textit{rr}, and \textit{c} refer to the left-left, right-right, and center views, respectively. Results include End-Point Error (EPE) and outlier ratios (errors $>3$px) across general, occluded, and out-of-frame regions."Occ" and "Oof" represent the occluded regions and the out-of-frame regions, respectively.}
\begin{tabular}{c|cccccc}
\hline
\multirow{3}{*}{\textbf{Method}} & \multicolumn{6}{c}{\textbf{KITTI 2012}} \\ \cline{2-7} 
 & \multicolumn{3}{c|}{\textbf{EPE}$\downarrow$} & \multicolumn{3}{c}{\textbf{$>$3px(\%)}$\downarrow$} \\ \cline{2-7} 
 & \textbf{All} & \textbf{Occ} & \multicolumn{1}{c|}{\textbf{Oof}} & \textbf{All} & \textbf{Occ} & \textbf{Oof} \\ \hline
Baseline & 1.44 & 5.01 & \multicolumn{1}{c|}{15.58} & 7.5 & 40.0 & 66.3 \\
+ \textit{ll} + \textit{rr} & 1.24 & 4.58 & \multicolumn{1}{c|}{10.65} & 6.0 & 37.7 & 53.4 \\
+ \textit{c} & 1.41 & 4.96 & \multicolumn{1}{c|}{14.83} & 6.8 & 40.1 & 65.6 \\
+ \textit{ll} + \textit{rr} +\textit{c} & 1.16 & 4.39 & \multicolumn{1}{c|}{9.77} & 5.81 & 36.1 & 51.4 \\ \hline
\end{tabular}
\label{tab:additional ablation results}
\end{table}

\subsubsection{Self-Supervised Stereo Matching}
\label{sec:additional_visualziation_results}
In addition to the ablation studies presented in the main paper, we further evaluated the impact of multi-baseline stereo images generated by our DMS on self-supervised stereo-matching performance using the KITTI 2012 dataset. Same as the main paper, we also use PASMNet ~\cite{PASMNet} as the baseline for self-supervised stereo-matching training.   

As detailed in Table~\ref{tab:additional ablation results}, the results show that adding \textit{ll} and \textit{rr} significantly reduces both EPE and outlier ratios in all regions compared to the baseline, particularly improving occluded and out-of-frame areas. Further incorporating the center view (\textit{c}) yields the best performance, achieving the lowest EPE and outlier percentages, demonstrating the effectiveness of multi-baseline integration in enhancing geometric consistency and depth estimation robustness.

\begin{figure}[!t]
    \centering
    \includegraphics[width=1.0\linewidth]{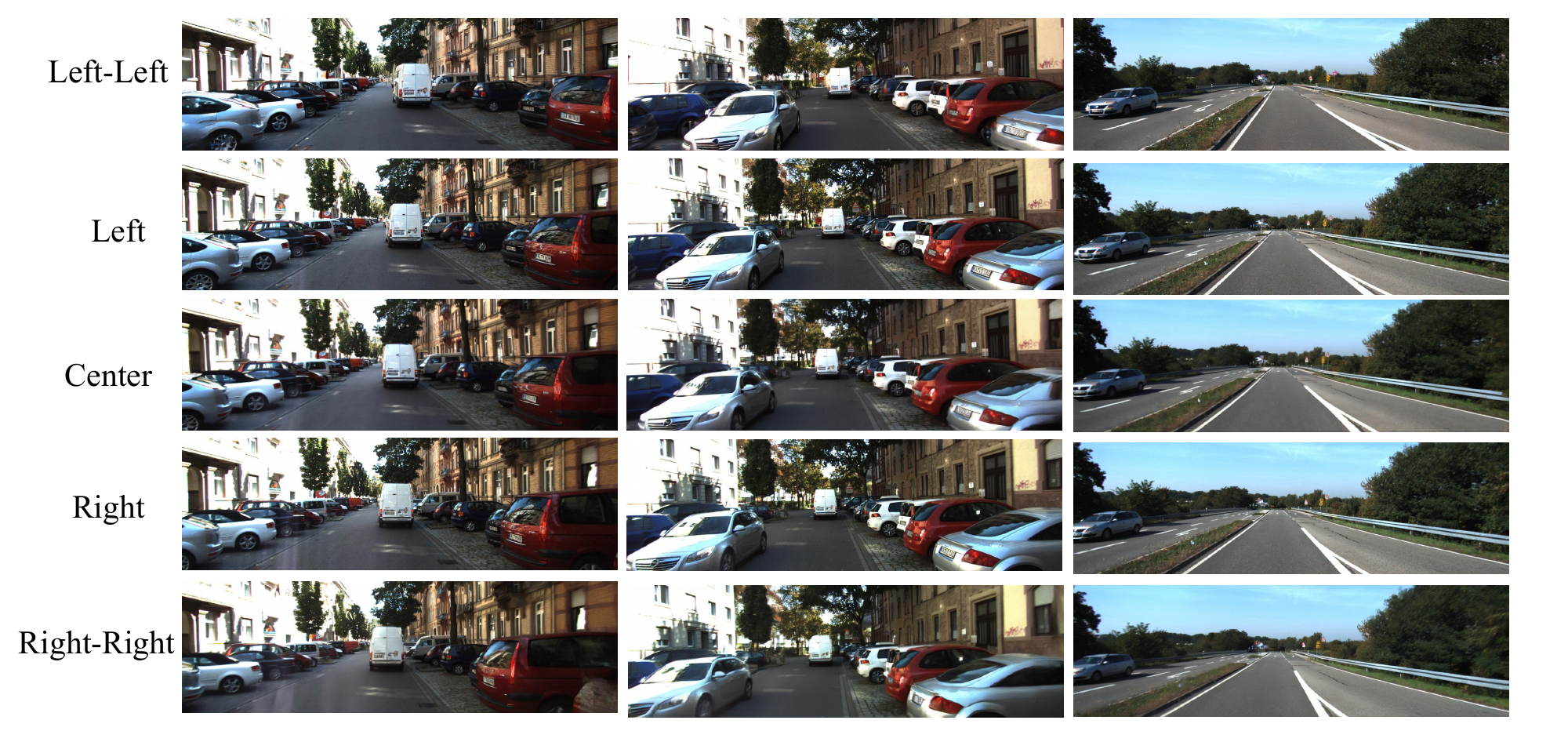}
    \caption{Multi-baseline stereo images generation using proposed DMS on the KITTI 2015 dataset.}

\label{fig:supp:KITTI_DMS}
\end{figure}

\subsection{Additional Visualization Results.}
\label{sec:supp:additional_vis_results}
\subsubsection{Multi-Baseline Stereo Image Generation Results Visualization.}

\begin{figure}[!t]
    \centering
    \includegraphics[width=1.0\linewidth]{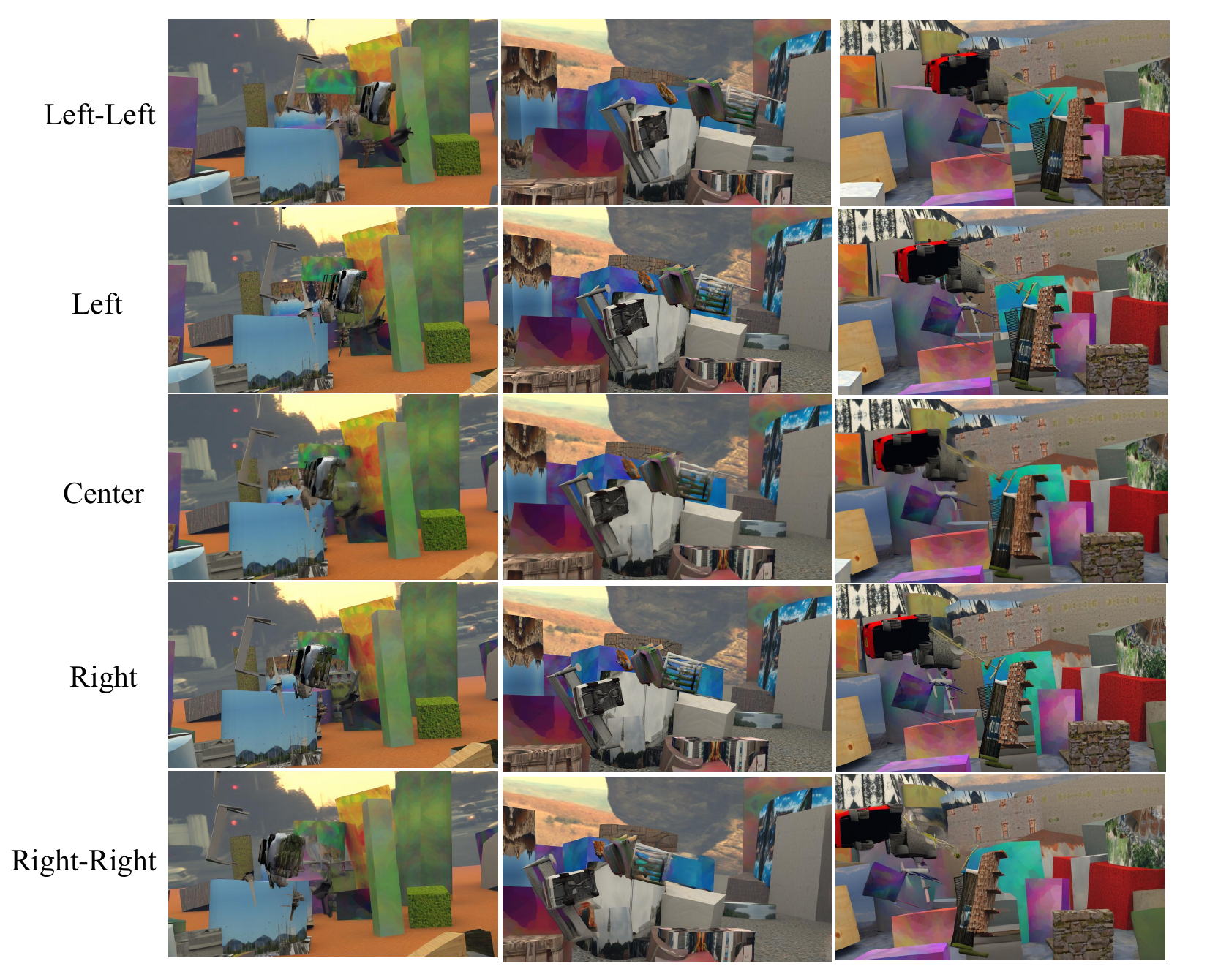}
    \caption{Multi-baseline stereo images generation using proposed DMS on the SceneFlow dataset.}
\label{fig:supp:SF_DMS}
\end{figure}

\begin{figure}[!t]
    \centering
    \includegraphics[width=1.0 \linewidth]{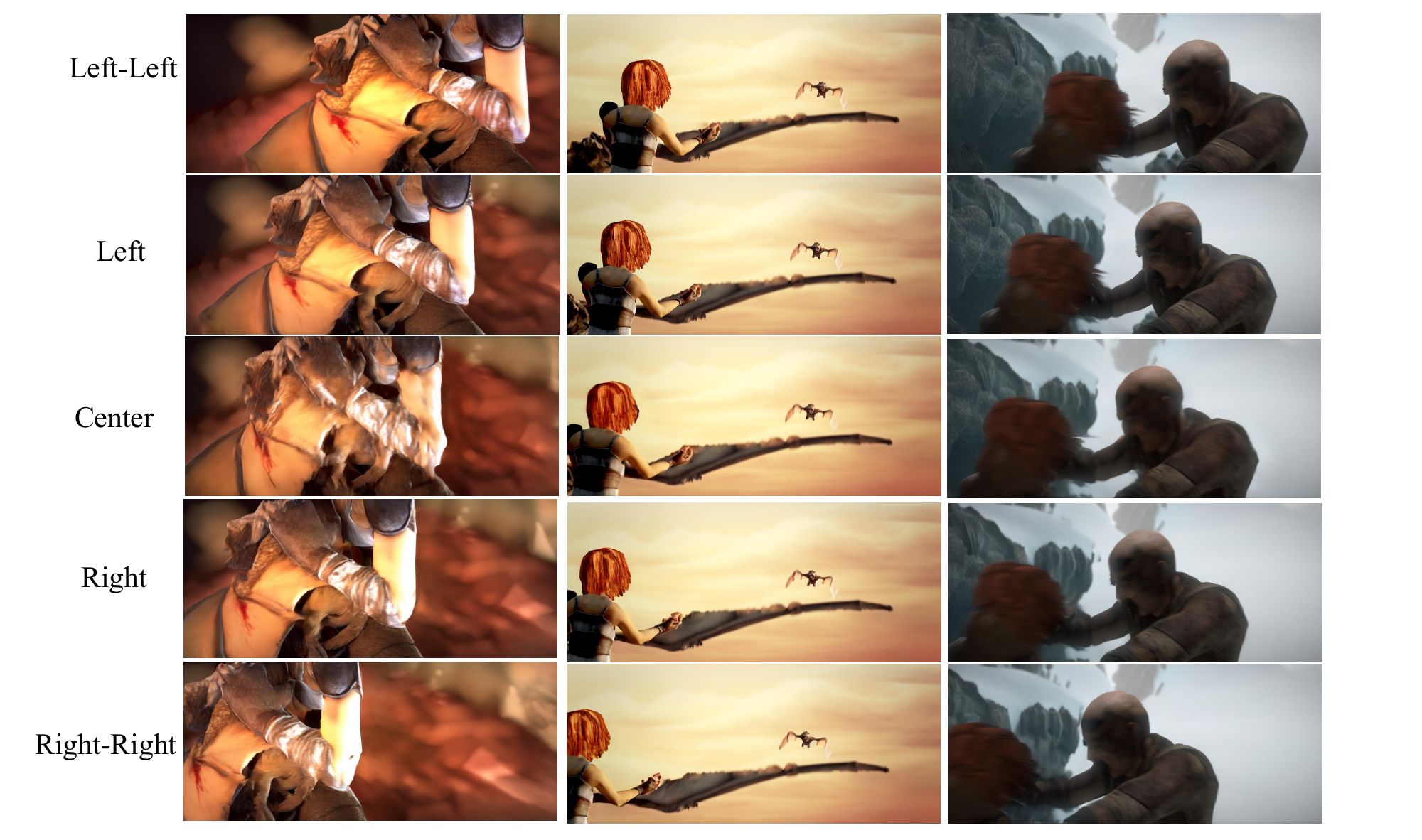}

    \caption{Multi-baseline stereo images generation using proposed DMS on the MPI-Sintel dataset.}

\label{fig:supp:mpi_DMS}  
\end{figure}

Figure~\ref{fig:supp:KITTI_DMS}, Figure~\ref{fig:supp:SF_DMS}, Figure~\ref{fig:supp:mpi_DMS}, and Figure~\ref{fig:supp:mpi_CARLA} showcase additional visualizations of the proposed DMS model applied to the SceneFlow, KITTI, MPI-Sintel, and CARLA datasets. These figures illustrate the model's capability to synthesize novel views along the epipolar line, leveraging directional prompts to extend stereo baselines. The results highlight the versatility and robustness of DMS in handling diverse scenarios, from highly controlled synthetic datasets to complex real-world environments, while maintaining geometric consistency.

{
    \small    \bibliographystyle{ieeenat_fullname}
    \bibliography{main}
}

\end{document}